%% file: main.tex
\input{preemble}

\title{\large\model~\ihydra: A Navigation World Action Model with \\Discrete Latent Planning and Continuous Flow-Matching Execution} %

\author{
Mohammad Nazeri\thanks{George Mason University, \texttt{\{mnazerir, acard, apokhrel, xiao\}@gmu.edu}} \And
Alexandyr Card\footnotemark[1] \,$^{\P}$ \And
Samira Huber\thanks{Kiel University, \texttt{\{samira.huber, ruben.hammele, sp\}@informatik.uni-kiel.de}} \,$^{\P}$ \And
Anuj Pokhrel\footnotemark[1] \,$^{\P}$ \And
Yujun Wang\thanks{Ludwig Maximilian University Munich, \texttt{yujun\_wang\_cn@hotmail.com}} \,$^{\P}$ \And
Ruben Hammele\footnotemark[2] \,$^{\P}$ \And
Daeun Song\thanks{Ewha Womans University, \texttt{songd@ewha.ac.kr}} \And
Sören Pirk\footnotemark[2] \And
Xuesu Xiao\footnotemark[1]%
}

\begin{document}

\maketitle
{
  \renewcommand{\thefootnote}{\P}
  \footnotetext{Equal contribution as second authors.}
}

\begin{figure}[H]
\vspace{-1cm}
  \setlength{\linewidth}{\textwidth}
  \setlength{\hsize}{\textwidth}
  \centering
    \includegraphics[width=1\textwidth]{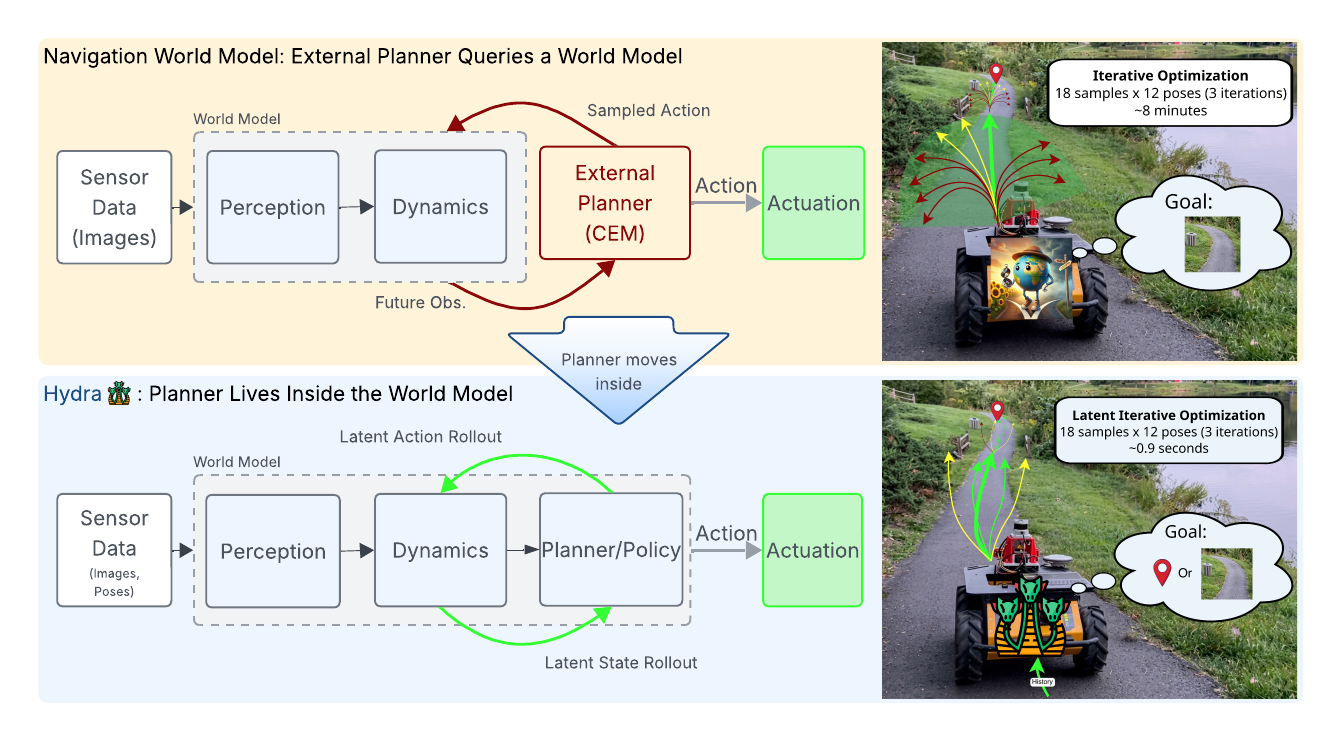}
    \caption{\textbf{Moving the planner into the model's own latent space.} \textbf{Top:} NWM uses an external planner to sample unconstrained trajectories, decoding them to pixels for evaluation (\red{red}), wasting compute on infeasible candidates. \textbf{Bottom:} \model~plans directly within its discrete latent manifold, then executes via continuous Flow Matching, achieving $>500\times$ faster planning on a robot.}
    \label{fig:teaser}
\end{figure}

\begin{abstract}
\input{contents/abstract}
\end{abstract}
\section{Introduction}\label{sec:intro}
\input{contents/introduction}

\section{Related Work}\label{sec:related_work}
\input{contents/related_work}

\section{Hydra}\label{sec:approach}
\input{contents/approach}

\section{Experiments}\label{sec:experiments}
\input{contents/experiments}

\input{contents/discussions}

\input{contents/acknowledgment}

\bibliographystyle{plain}
\bibliography{bibliography}

\input{contents/appendix}

\end{document}

%% file: preemble.tex
\documentclass{article}
\usepackage{arxiv}

\usepackage[utf8]{inputenc} 
\usepackage[T1]{fontenc}    
\usepackage{booktabs}       
\usepackage{amsfonts}       
\usepackage{amssymb}
\usepackage{dsfont}
\usepackage{nicefrac}       
\usepackage{microtype}      
\usepackage{graphicx}
\usepackage{algorithm}
\usepackage{algpseudocode}
\usepackage{multirow}
\usepackage{wrapfig}
\usepackage{colortbl}
\usepackage{enumitem}
\usepackage{amsmath}
\usepackage{natbib}
\usepackage{flushend}
\usepackage{nicematrix}     
\usepackage{adjustbox}
\usepackage{fontawesome5}
\usepackage{glossaries}

\usepackage{eccvabbrv}
\usepackage{gradient-text}
\usepackage[dvipsnames]{xcolor}
\usepackage[accsupp]{axessibility}  
\usepackage{hyperref}
\usepackage{subfig} 
\usepackage{gensymb} 
\usepackage{color, soul} 

\hypersetup{
    colorlinks=true,
    citecolor=blue, 
    linkcolor=blue,
    filecolor=magenta,      
    urlcolor=cyan,
    pdftitle={HYDRA},
    }

\definecolor{HydraBlue}{HTML}{1A4B84}   
\definecolor{FlowTeal}{HTML}{00A9A5}   
\definecolor{ActionCoral}{HTML}{F05D5E} 

\definecolor{LatentPurple}{HTML}{6B4E71} 
\definecolor{SlateGray}{HTML}{4A5568}    
\definecolor{LightBlueBg}{HTML}{EBF4FA}  
\definecolor{DarkRed}{HTML}{8B0707} 

\definecolor{green}{RGB}{11,155,13}
\definecolor{blue}{RGB}{15, 71, 109}
\definecolor{lightblue}{RGB}{15, 25, 180}

\newcommand{\red}[1]{{\textcolor{DarkRed}{#1}}}

\newcommand{\model}{\textcolor{HydraBlue}{Hydra}}

\DeclareRobustCommand{\ihydra}{%
  \begingroup\normalfont
  \includegraphics[height=\fontcharht\font`\B]{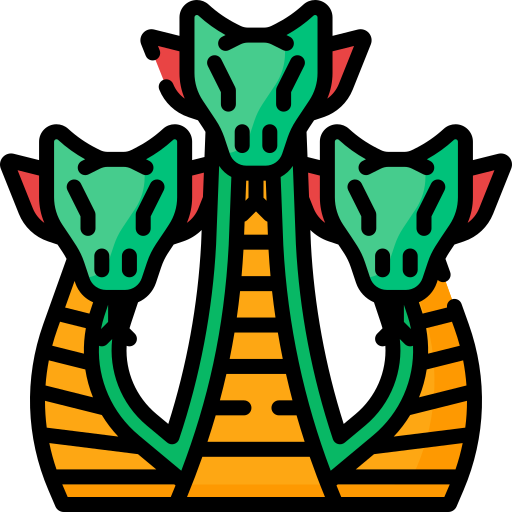}%
  \endgroup
}

\hypersetup{
pdftitle={Hydra},
pdfkeywords={Navigation, World Action Model},
}

%% file: contents/abstract.tex
World models let robots imagine possible futures, but exploiting this capability for real-time planning is bottlenecked by a \emph{representation misalignment}: generative models and planners operate on decoupled manifolds, requiring computationally expensive decoding of every candidate back to the high-dimensional observation space for evaluation. In this paper, we present \model, a discrete World Action Model that tackles this by establishing a unified latent manifold over visual states, physical poses, and control actions. By compressing this manifold through modality-specific Vector-Quantized bottlenecks, \model~yields discrete vocabularies of kinodynamic intents and visual states. This enables \textit{Discrete Latent Planning} (DLP), where candidates are sampled directly from the shared manifold and ranked by a Kinematic-Perceptual Cost within the discrete latent space. To bridge discrete planning with the continuous commands required for physical actuation, \model~pairs DLP with conditional Flow Matching to map selected intents to smooth execution trajectories. Evaluated on two physical robotic platforms, \model~outperforms state-of-the-art navigation world models in goal-directed planning, while matching or exceeding the closed-loop execution capabilities of leading reactive navigation policies\footnote{Model weights and deployment code are available at the project website: https://robotixx.github.io/hydra.}.

%% file: contents/introduction.tex
Visual foundation policies~\citep{shah2023gnm, shah2023vint, sridhar2023nomad} successfully map high-dimensional observations to motor commands for short-horizon, image-goal navigation, where the goal is visually reachable. However, their reactive nature limits their ability to plan ahead: lacking any mechanism to reason beyond immediate visual cues, making them prone to wandering off when the goal is not observable. Furthermore, the lack of an explicit cost formulation restricts their ability to adapt without policy retraining.

Recent approaches use World Models (WMs)~\citep{ha2018world, bar2025navigation} to predict future states conditioned on candidate actions from external planners like CEM~\citep{rubinstein1997optimization} or MPPI~\citep{williams2015model}, giving the agent the foresight reactive policies lack. However, this introduces a fundamental representation misalignment: planners search an unbounded, continuous space decoupled from the model's latent manifold, requiring computationally expensive decoding to raw observations to evaluate candidates.

To bridge this gap, we introduce \model, a World Action Model (WAM) that embeds the sampling-based planner and candidate trajectory evaluator entirely within a unified latent manifold (Fig.~\ref{fig:teaser}). Because sampling now draws directly from this joint manifold, candidate actions are informed by the model's own understanding of the observation rather than proposed blindly; because the manifold is bounded, search remains manageable across longer horizons; and because evaluation happens within this shared space, candidates are scored without ever decoding back to pixels.

To achieve this, \model~feeds its future predictions through modality-specific Vector Quantized (VQ)~\citep{van2017neural} codebooks, compressing the unbounded, continuous action space into a discrete manifold of \textit{intents} -- macro-level kinodynamic primitives, such as a sweeping left turn, drawn from a small vocabulary learned from the training data. This substitution changes what planning actually is: rather than optimizing over a continuous space in which most candidate paths are physically implausible, \model~searches over a finite set of behaviors already known to be plausible, with the code retrieved at each step being the one whose embedding lies closest to \model's embedding of the observation history, rather than an arbitrary vocabulary entry. 

We term this search \textit{Discrete Latent Planning} (DLP). As Fig.~\ref{fig:teaser} illustrates, this changes not only how many candidates are considered but what kind: NWM's continuous sampler (red) spreads trajectories indiscriminately, evaluating every one to discover which are bad; \model's candidates (green) are drawn only from codes the model has already learned to associate with the observed scene, so implausible trajectories are rarely sampled to begin with.

Because DLP operates in the discrete latent space, candidate evaluation is efficient. \model~utilizes a Kinematic-Perceptual Cost ($\mathcal{J}_{\text{KPC}}$) where Shannon entropy represents predictive uncertainty, and out-of-distribution (OOD) visual quantization error acts as a direct proxy for anomalous states, such as impending collisions, without ever decoding to pixels. The selected discrete intent is subsequently passed to continuous Flow Matching ODE solvers, integrating the macro-level intent into the continuous motor commands required for smooth physical execution. Ultimately, \model~demonstrates capability as both a planner and a reactive path follower, matching the execution precision of reactive navigation policies while outperforming baseline decoupled planner and World Models.

In summary, our main contributions are: 
1) \textbf{\model:} A multimodal World Action Model that unifies visual, kinodynamic, and action modalities into a single latent space, moving the planner, the search space, and cost evaluation into the model's unified latent manifold rather than leaving it external and decoupled.
2) \textbf{Discrete Latent Planning:} An iterative latent search, guided by a multi-objective Kinematic-Perceptual Cost, that avoids costly decoding to observations with search and safety evaluation within a discrete manifold.
3) \textbf{Real-World Deployment:} \model~functions as both a planner and a reactive path follower, outperforming NWM~\citep{bar2025navigation} in planning while matching the execution precision of dedicated reactive navigation policies.

%% file: contents/related_work.tex
To show the contributions of \model, we review prior literature across three dimensions: \textit{Reactive Navigation Policies}, which lack the generative foresight required for planning; \textit{Generative World Models}, which predict but often lack explicit Euclidean grounding; \textit{Planning with a Model}, where decoupled sampling imposes computational bottlenecks that \model's discrete latent planning avoids.

\textbf{Reactive Navigation Policies:}
Driven by large-scale heterogeneous datasets, recent efforts have produced capable end-to-end navigation policies~\citep{shah2023gnm, shah2023vint, sridhar2023nomad, liu2025citywalker}, including diffusion-based approaches that model multimodal action distributions~\citep{chi2025diffusion, nguyen2025pixel, zheng2025diffusion, pan2025adonoisingdispellingmyths}. These architectures are fundamentally reactive, compressing navigation into a direct observation-to-action mapping without an internal model of future consequences. Because they are open-loop with respect to the future environment, they cannot evaluate or prune candidate trajectories before execution, making them myopic and incapable of planning.

\textbf{Generative World Models:}
To endow robots with foresight, WMs~\citep{ha2018world, wu2023daydreamer, janner2022planninga, yu2023scaling, nvidia2025cosmos, park2026dreamflow, zhu2025adljepa, maes2026leworldmodel, kim2026cosmos, nvidia2026cosmos} simulate the consequences of actions, evolving from early predict-then-act pipelines~\cite{du2023learning, wen2024vidman, bharadhwaj2025gen2act} toward unified diffusion~\citep{sohl-dickstein2015deep, rombach2022highresolution, hoogeboom2022autoregressive} or Flow Matching~\citep{liu2022flow, lipman2023flow, xing2025goalflow} architectures. Visual-action models~\citep{hu2023gaia1, russell2025gaia2, bar2025navigation, videoworldsimulators2024} and large-scale unified backbones~\citep{li2025unified, zhu2025unified, ali2025world, octomodelteam2024octo, mereu2025generative, worldlabs_2025, team2026gigabrain, team2026advancing} provide tighter coupling but typically at billions of parameters for general-purpose manipulation; DreamZero~\citep{ye2026world}, for instance, uses a 14B-parameter WAM as a zero-shot policy on physical manipulators, performing no search since rollout-based planning at this scale is computationally prohibitive. NavWM~\citep{mei2026navwm} targets navigation with a Mamba backbone, but discretizes only candidate goal locations via a K-means anchor codebook; so evaluation still requires generated visual rollouts rather than a decode-free latent score. \model~targets navigation instead, at 143M parameters, sized for real-time search.

\textbf{Planning with a Model:}
Sampling-based planning approaches such as MPPI or CEM~\citep{zhou2024dinowm, bar2025navigation, maes2026leworldmodel} use World Models as the dynamics forward model; for navigation, NWM (via CEM) and VertiFormer~\citep{nazeri2025vertiformer} (via MPPI) are our closest baselines. Both search an unconstrained continuous action space, but the consequences differ by architecture: lacking a native action head, NWM must decode each candidate and evaluate it with a perceptual metric such as LPIPS~\citep{zhang2018unreasonable}, which is prone to texture-matching illusions rather than measuring true spatial distance; VertiFormer instead retains a direct geometric cost from its action head, but its unbounded search space still admits implausible, collision-prone candidates absent a structural prior.
Latent-space planning~\citep{fang2023generalization, zheng2025flare, su2026world, zhou2024dinowm, sobal2025learning, assran2025vjepa, sun2026vla} avoids pixel rendering, but JEPA-style models~\citep{assran2023selfsupervised, bardes2023mcjepa, zhu2025adljepa, assran2025vjepa, balestriero2025lejepa} still evaluate continuous action manifolds, inheriting the same sample inefficiency. Discretizing the latent space is not itself new; DreamerV3~\citep{hafner2023mastering} and IRIS~\citep{micheli2023transformers} do so, but neither couples discretization to a candidate-ranking cost function nor targets physical, real-time robot planning. WorldPlanner~\citep{khorrambakht2025worldplanner} discretizes the Monte Carlo Tree Search rather than the underlying representation, so actions and states remain continuous and pixel-valued, requiring decode-then-evaluate and taking five minutes on average to plan. LeFlow~\citep{huang2026leflow}, applied to toy problems, amortizes search by learning a rectified-flow prior that generates a goal-conditioned latent path in one forward pass; built on a JEPA backbone with no visual decoder, it avoids decoding architecturally, but its representation remains continuous and unconstrained. \model~discretizes the representation itself: kinodynamic intents and visual states are drawn from finite, learned vocabularies, so search and cost evaluation happen entirely within this discrete space, on a physical robot rather than in simulation.

%% file: contents/approach.tex
\input{contents/problem_formulation}

\subsection{\model}
\label{sec:hydra_model}

In this section, we discuss the three primary mechanisms of \model: unified early-fusion encoding, discrete decoding, and multimodal flow matching. Figure~\ref{fig:arch_baselines} shows \model's architecture compared with previous approaches. In the Appendix, Table~\ref{tab:comparison} compares \model's modalities with existing approaches.

\textbf{Unified Multimodal Encoding via Early Fusion.}
Visual latents $z_i \in \mathbb{R}^{D_z}$ are extracted via a frozen VAE~\citep{blattmann2023stable}, and low-dimensional actions and poses undergo Fourier Feature Mapping~\citep{tancik2020fourier, zhou2023nerf} to capture sharp control transitions.
To avoid computationally heavy cross-attention between vision and kinodynamics, \model~adopts an early-fusion unified encoding strategy akin to VertiFormer~\citep{nazeri2025vertiformer}. More details are provided in Appendix~\ref{app:architecture}.

We concatenate this unified history stream with $H$ learnable context tokens, where $H$ is the forecasting horizon. The Transformer Encoder processes this joint sequence and outputs a fixed-size set of context tokens, $\text{ctx}_{t+1:t+1+H}$. These tokens compress the ``where, what, and how'' of the robot's history.

\textbf{The Discrete Bottleneck.}
The Transformer Decoder cross-attends the context tokens $\text{ctx}_{t+1:t+1+H}$ with the intended future poses ($p_{t+1:t+1+H}$) or actions ($a_{t:t+H}$), utilizing the same missing-modality infilling as the encoder. This yields $H$ ``intent'' tokens ($z_e$), which are subsequently fed to modality-specific VQ codebooks~\citep{van2017neural, lee2022autoregressive, zargarbashi2026vq}, one per modality. 
To balance high-fidelity spatial forecasting with the strict latency constraints of closed-loop control, we employ \textit{asymmetric vocabulary quantization}. We allocate a large codebook ($|\mathcal{V}_{i}| = 2048$) to the visual generative head, while strictly constraining the kinodynamic intents ($|\mathcal{V}_{p}| = 64, |\mathcal{V}_{a}| = 64$). This asymmetry preserves visual representational capacity while explicitly enabling the fast retrieval speeds required for DLP. To stabilize training, we applied deterministic anchoring; see App.~\ref{sec:learning_perspective} for details.

\textbf{Multimodal Generative Flow Matching.}
For each modality $m \in \{\text{img,pose,act}\}$, we define a probability path from a source Gaussian distribution $\psi_0^{(m)} \sim \mathcal{N}(0, I)$ to the ground-truth target $\psi_1^{(m)}$. At any continuous integration time $s \in [0, 1]$, the intermediate noisy state is $\psi_s^{(m)} = (1 - s)\psi_0^{(m)} + s \psi_1^{(m)}$. \model~employs three flow decoders~\citep{liu2022flow}, one per modality, denoted $v_{\theta}^{(m)}$, to predict the target velocity field for images, poses, and actions.

Each discrete token $c^{(m)}$, the index selected from the VQ codebook, is used to retrieve its corresponding embedding $e_c^{(m)}$. This embedding, along with the time step $s$, conditions the flow decoders via DiT blocks~\citep{peebles2023scalable}. The network is optimized end-to-end via the vector field loss, summed over modalities:
\begin{equation}
\mathcal{L}_{flow} = \sum_{m} \mathbb{E}_{s, \psi_0^{(m)}, \psi_1^{(m)}} \left[ | v_{\theta}^{(m)}(\psi_s^{(m)}, s, e_c^{(m)}) - (\psi_1^{(m)} - \psi_0^{(m)}) |_2^2 \right].
\nonumber
\end{equation}
 
Please refer to Appendix~\ref{sec:learning_perspective} for more details about the design choices.

\input{contents/planning}

%% file: contents/problem_formulation.tex
\begin{figure}[!t]
    \centering
    \includegraphics[width=0.95\linewidth]{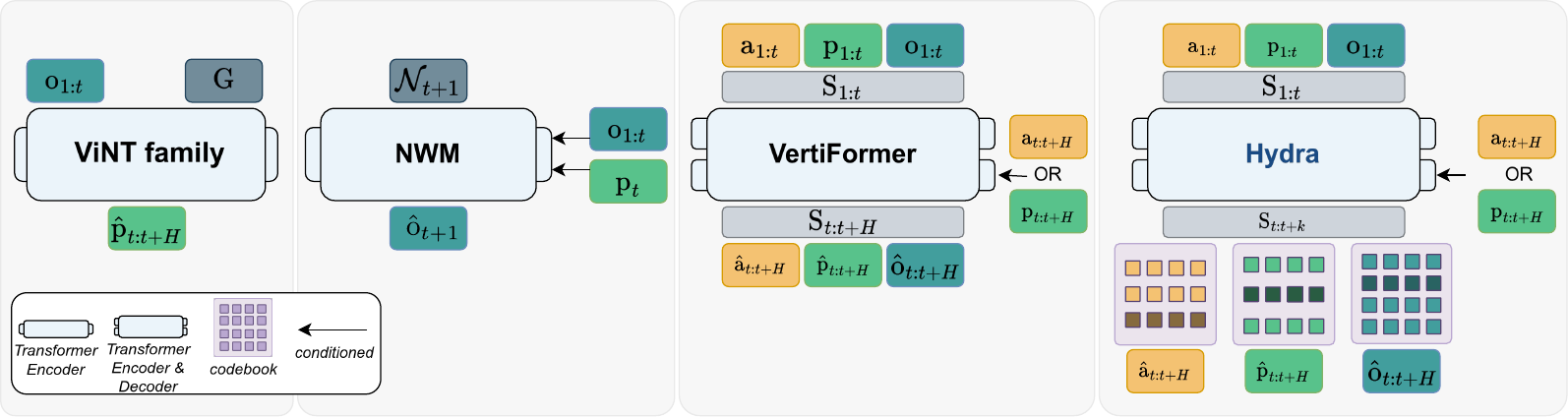}
    \caption{Comparison of \model~architecture with Visual Navigation Transformer (ViNT), Navigation World Model (NWM), and VertiFormer in terms of input modalities, conditioning, and the model output. \model~builds on top of VertiFormer and discretizes the output into codebooks to enable the model to search the discrete latent space for a plausible path.}%
    \label{fig:arch_baselines}%
\end{figure}

This section presents \model~in four parts. We first formalize autonomous navigation through the lens of three paradigms (see Fig.~\ref{fig:arch_baselines}), situating \model's unified planning relative to reactive policies and decoupled world models (Sec.~\ref{sec:problem_formulation}). 
We then detail \model's architecture, specifically the discrete codebook 
(Sec.~\ref{sec:hydra_model}). Next, we explain how \model~conducts Discrete Latent Planning within these codebooks, searching and ranking trajectories without pixel-level decoding (Sec.~\ref{sec:planning}). Finally, we describe \model's local path following strategies when guided by an external global planner (Sec.~\ref{sec:path_following}).

\subsection{Problem Formulation: Navigation Paradigms}
\label{sec:problem_formulation}

Autonomous navigation requires predicting an action sequence to achieve a goal in a partially observable environment. At timestep $t$, the model receives a multimodal observation $O_t = \{I_t, p_t, a_{t-1}\}$, comprising ego-centric vision $I_t \in \mathbb{R}^{3 \times H \times W}$, odometric pose $p_t = (x_t, y_t, \theta_t)$, and the prior motor command $a_{t-1} = (v, \omega)$.

The agent maintains a context history $\mathcal{H}_t = \{O_{t-K+1}, \dots, O_t\}$ over a window $K$. Based on how models map this history to actions, we categorize navigation models into three formulations. While we detail \emph{Reactive Policies} and \emph{Decoupled Planning} in App.~\ref{app:problem_formulation}, the third paradigm, \emph{Unified Planning}, forms the basis of \model~and is described below.

\textbf{World Action Model Formulation (Unified Planning):}
\model~formulates planning so that the planner and the model share the same embedding: visual states, poses, and actions are all sampled and evaluated within a single latent manifold, as opposed to a decoupled model where the planner sends actions to and receives images from the model.

Given the observation history $\mathcal{H}_t$ and a sampled sequence of future actions $a_{t:t+H-1}$ or poses $p_{t+1:t+H}$, the model $P_{WA}$ generates the expected future observations:

\begin{equation}
    (\hat{I}_{t+1:t+H}, \hat{y}) \sim P_{WA}(\cdot \mid \mathcal{H}_t, x)
    \label{eq:hydra_formula}
\end{equation}
where
\begin{equation*}
    (\hat{y}, x) = 
    \begin{cases} 
        (\hat{p}_{t+1:t+H}, a_{t:t+H-1}) & \text{if forward kinodynamics}, \\
        (\hat{a}_{t:t+H-1}, p_{t+1:t+H}) & \text{if inverse kinodynamics}.
    \end{cases}
\end{equation*}

%% file: contents/planning.tex
\subsection{Planning in the Discrete Latent Space}
\label{sec:planning}

\begin{figure}[t]
    \centering
    \includegraphics[width=0.95\linewidth]{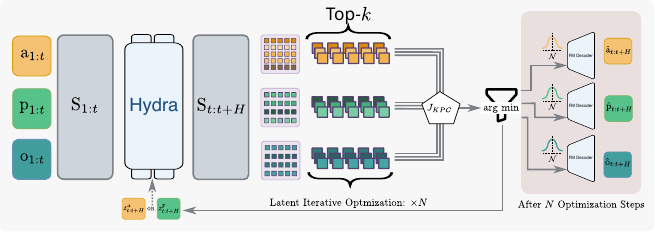}
    \caption{Initial intended future pose/action tokens are generated unconditionally, then Gaussian noise is added to become the exploratory seed `intents'. Then these future intents go through VQ codebooks, and we sample the top-$k$ codes. After calculating the KPC cost, the lowest cost is chosen to seed the next iteration of searching. The best trajectory is decoded using flow matching heads to continuous robot poses and actions at the end for robot execution.}%
    \label{fig:planning}%
\end{figure}

In decoupled planning and world models, candidate actions are proposed without any knowledge of the observed scene, and evaluating them requires decoding back to the raw observation (\eg, pixels) rather than comparing candidates directly in latent space. \model~closes this gap by coupling the planner to the model itself, introducing \textit{Discrete Latent Planning} (DLP) (\eqref{eq:hydra_formula}) in place of the decoupled formulation in \eqref{eq:wm_dynamics}. Instead of searching blindly in a continuous space, \model~operates explicitly over the learned manifold defined by the discrete modality codebooks $\mathcal{V}_{i}$, $\mathcal{V}_{p}$ and $\mathcal{V}_{a}$.

\textbf{Unbiased Initialization and Gumbel-Max Search.}
In the absence of an explicit future intent, \model~initializes its search using an \textit{Unbiased Expert Prior} by querying the model with masked future states.
Using this unbiased seed, \model~generates a diverse batch of candidate paths by adding Gaussian noise. Furthermore, by applying exponentially smoothed Gumbel noise to the discrete codebook logits, \model~injects path perturbations while preserving kinodynamic smoothness across time steps.

\textbf{Latent Optimization.}
Once candidates are retrieved, they must be ranked. Previous visual planning models such as NWM  
used perceptual similarity such as LPIPS~\citep{zhang2018unreasonable}, which incur massive computational overhead due to the need to decode the latents to pixels for computation.
To avoid this, \model~evaluates candidates within the discrete manifold using a multi-objective cost function, defined for a candidate sequence $\tau$ as:

\begin{equation}
\label{eq:kpc}
\mathcal{J}_{\text{KPC}}(\tau) = 
\underbrace{\mathcal{C}_{\text{geo}}}_{\text{Geometric Goal}} + 
\sum_{t=1}^{T} \gamma^t \Bigg( 
\underbrace{\lambda_{\text{img}}\mathcal{C}_{\text{img}}^{(t)}}_{\text{Visual Goal}} + 
\underbrace{\lambda_{\text{prior}}\mathcal{C}_{\text{prior}}^{(t)}}_{\text{Kinodynamic Prior}} + 
\underbrace{\lambda_{\text{ent}}\mathcal{C}_{\text{ent}}^{(t)} + \lambda_{\text{vq}}\mathcal{C}_{\text{vq}}^{(t)}}_{\text{Perceptual Safety}} 
\Bigg)
\end{equation}

where $\gamma \in (0, 1)$ is a temporal discount factor that decays the weight of distant future states to account for compounding generative errors.

\begin{itemize}[leftmargin=*, nosep]
    \item \textbf{Geometric and Semantic Tracking ($\mathcal{C}_{\text{geo}}, \mathcal{C}_{\text{img}}$):} both are evaluated as $L_2$ deviations from the goal,  $\mathcal{C}_{\text{geo}}$ from the local goal coordinates and $\mathcal{C}_{\text{img}}$ from the visual goal latent embedding. A direct path minimizes this cost, but requires regularization to survive occlusions.
    \item \textbf{Unbiased Expert Prior ($\mathcal{C}_{\text{prior}}$):} penalizes deviations from collision-free demonstrations via the Negative Log-Likelihood (NLL) of sampled tokens.
    \item \textbf{Visual Predictive Entropy ($\mathcal{C}_{\text{ent}}$):} computes the Shannon entropy of the predicted visual token distribution, penalizing trajectories that enter ambiguous or unobservable regions.
    \item \textbf{Manifold Rejection ($\mathcal{C}_{\text{vq}}$):} computes the quantization error between the continuous sampled intent and the visual codebook $\mathcal{V}_{i}$. OOD states (\eg, driving into an occlusion) fall off the learned manifold of safe navigation, yielding high quantization error and allowing rejection without invoking the visual decoder.
\end{itemize}

Note that all KPC terms are calculated in the latent space except $\mathcal{C}_{\text{geo}}$, which is decoded using the deterministic head to compute the Euclidean distance to the goal, a negligible computational cost.
By evaluating $\mathcal{J}_{\text{KPC}}(\tau)$ on the intent tokens, iterative refinement runs only the transformer decoder. Flow matching is deferred, called only once to decode the chosen trajectory ($N=10$ NFEs for kinodynamic; $N=50$ for visual). This latent formulation drives a $>500\times$ reduction in planning time relative to NWM (Table~\ref{tab:global_planning}). Finally, the discrete tokens of this trajectory are decoded into local spatial deltas and fused via Action Chunking~\citep{zhao2023learning} for smooth closed-loop execution.

\subsection{Path Following via Global Guidance}
\label{sec:path_following}
While DLP enables standalone planning, long-horizon navigation often requires tracking global subgoals $g$ provided by a planner. \model~acts as a reactive policy that minimizes deviation from $g$ within obstacle-free regions. We propose two strategies for path following:

\textbf{Classifier-Free Guidance in Kinodynamic Space.}
We adapt CFG~\citep{ho2022classifier} to blend structural safety with goal-directed tracking. We compute an unconditional prior $z_{\emptyset}$ and a geometric intent $z_{g}$, interpolating them:

\begin{equation}
\tilde{z}_{g} = z_{\emptyset} + \omega \cdot (z_{g} - z_{\emptyset}),
\end{equation}

with guidance scale $\omega$. However, high $\omega$ risks \textit{manifold deviation}, pushing $\tilde{z}_{g}$ outside the discrete VQ codebook bounds, causing the decoder to hallucinate physically impossible dynamics (App.~\ref{app:cfg}).

\textbf{Sub-Goal Sampling.}
Alternatively, we generate candidate trajectories by perturbing the continuous intent token with Gaussian noise. We decode these into a candidate set and select the trajectory minimizing the tracking cost:
\begin{equation}
  \mathcal{J}_{\text{track}}(\tau) = \| \tau_T - g \|_2 + \frac{1}{2T} \sum_{t=1}^{T} \| \tau_t - g \|_2 ,
  \label{eq:subgoal}
\end{equation}

where $T$ is the trajectory length. Because candidates are decoded strictly from the expert prior vocabulary, they inherently satisfy learned dynamic constraints and local obstacle avoidance (App.~\ref{app:sampling}).

%% file: contents/experiments.tex
The empirical evaluation of \model~is designed to answer three core questions: (1) \textbf{Generative Grounding:} does \model's parameter-efficient, early-fusion architecture maintain competitive visual predictive fidelity against continuous baselines like NWM over long horizons? (2) \textbf{Real-Time Planning Efficiency:} does coupling the planner to the model's latent manifold and evaluating candidates without decoding to pixels, bypass the computational bottlenecks of decoupled sampling and enable real-time deployment? (3) \textbf{Path-Following Accuracy:} as a local path follower, does \model~match reactive baselines (\eg, GNM, ViNT, NoMaD) without sacrificing observation-to-action behavior? Qualitative results (App.~\ref{app:rviz_branching}), pose codebook analysis (App.~\ref{sec:codebook_analysis}), and ablations validating the KPC terms (App.~\ref{sec:ablations} and \ref{sec:manifold_rejection_eval}) are provided in the Appendix.

\subsection{Generative Grounding: Long-Horizon Temporal Consistency}
\label{sec:video_comparison}

To evaluate whether \model's early fusion and discrete bottleneck preserve temporal grounding and video generation quality, following NWM, we autoregressively generate 100 video clips of 16 seconds (at 4 FPS) using the same held-out subset of the RECON dataset~\citep{shah2021rapid}. We benchmark against the base variant of NWM (NWM/B, 200M parameters DiT) and VertiFormer (a deterministic baseline), all evaluated under \textit{teacher-forcing} (ground-truth actions at each step) for a fair comparison of the visual generation.

\begin{wraptable}{R}{0.5\textwidth}
\centering
\caption{Comparison of Video Synthesis Quality. 100 clips of 16-seconds generated at 4 FPS.
}
\label{tab:fvd}
\footnotesize
\setlength{\tabcolsep}{3pt}
\begin{tabular}{l cc}
\toprule
\textbf{Model} & FVD $\downarrow$ & Time To Generate \\
\midrule
NWM/B & 22.10 $\pm$ 10.71 & 3.5h\\
VertiFormer & 81.03 $\pm$ 12.50 & 2m\\
\rowcolor{LightBlueBg} \textbf{\model} & \textbf{19.15} $\pm$ 5.11 & 4m\\
\bottomrule
\end{tabular}
\end{wraptable}

As shown in Table~\ref{tab:fvd}, \model~slightly outperforms NWM in Fréchet Video Distance (FVD), which measures temporal coherence and action adherence. However, both show high variance, depending on whether the robot moves straight. NWM suffers from compounding errors and visibly decouples from the commanded actions after approximately 8 seconds (Appendix~\ref{app:qualitative}), while \model's multimodal manifold enforces temporal grounding across the full 16-second horizon. 
We report single-step predictive fidelity (DreamSim, LPIPS, PSNR) in Appendix~\ref{app:image_quality}.

\subsection{Planning: Real-Time Feasibility via Discrete Search}
\label{sec:planning_exp}

While recent literature benchmarks generative planners using static offline datasets, we argue that this methodology is fundamentally misaligned with the objectives of planning. Offline dataset evaluations penalize any deviation from the human demonstration. If a planner discovers an alternative, physically valid, and collision-free trajectory, static evaluations incorrectly compute a high geometric penalty. True exploratory planning and obstacle avoidance should therefore be evaluated through closed-loop physical deployment. We explain the physical evaluation protocols, including robot type, scenario initialization, and hardware configurations for both planning and local path following in App.~\ref{app:experiments_setup}.

\textbf{Baselines.} We benchmark \model~against VertiFormer~\citep{nazeri2025vertiformer}, NWM/B, and Hydra-NoVQ, a variant of \model~that retains Flow Matching heads but removes VQ discretization, isolating the contribution of discrete search from Flow Matching. VertiFormer shares \model's early-fusion design but is deterministic, with neither VQ discretization nor Flow Matching, relying on MPPI sampling. All models use the SD's VAE~\citep{blattmann2023stable} with 80M frozen parameters. Without a discrete latent representation, neither baseline can leverage our $\mathcal{C}_{\text{prior}}$, $\mathcal{C}_{\text{ent}}$, and $\mathcal{C}_{\text{vq}}$ cost terms. Computing $\mathcal{C}_{\text{prior}}$ and $\mathcal{C}_{\text{ent}}$ requires a normalized probability distribution; over a finite codebook, this normalization is a tractable sum, whereas over a continuous action space it requires integrating a density over an unbounded domain, which is intractable. $\mathcal{C}_{\text{vq}}$ is unavailable because it measures distance to a learned codebook entry, and no such codebook exists in a continuous representation. Therefore, baselines are restricted to $\mathcal{C}_{\text{geo}}$ and $\mathcal{C}_{\text{img}}$ cost terms.

\input{contents/planning_comparison_table}

\textbf{The Failure of External Planners.} As shown in Table~\ref{tab:global_planning}, NWM suffered catastrophic latency under identical hyperparameters (3 iterations, 18 samples, length 12).
Due to the lack of a native action head, NWM's one CEM search took over 500 seconds, as it relied on a compute-intensive image goal comparison, which requires each candidate to be decoded to pixels.

VertiFormer and Hydra-NoVQ instead use a direct geometric cost $\mathcal{C}_{\text{geo}}$ from the predicted pose, letting them search orders of magnitude faster ($\sim$0.7, $\sim$1.2s) despite their unbounded space, but both suffer \textit{sample inefficiency}: without a structural prior bounding the search space, they generate collision trajectories, yielding a 0\% Success Rate across obstacle scenarios. Because Hydra-NoVQ retains \model's Flow Matching heads and differs only in the absence of VQ discretization, this isolates the source of \model's obstacle-avoidance capability as the discretization of the search space and the $\mathcal{C}_{\text{ent}}$, $\mathcal{C}_{\text{vq}}$, and $\mathcal{C}_{\text{prior}}$ terms it enables.

\textbf{Real-Time DLP.} Although our code is not optimized
to leverage quantization or kernel fusion on GPU, by restricting the search space to a discrete manifold of kinodynamic primitives directly in latent space without decoding, \model~plans in $\sim$0.9 seconds, a $\sim$500$\times$ speedup over NWM, while achieving a 10/10 success rate in unobstructed environments and 8/10 in both occluded and blind-corner scenarios.

\subsection{Local Path Following: \model~vs. Reactive Policies}
\label{sec:local_following_eval}

We next evaluate \model~as a \emph{local path follower} that tracks waypoints from an external guidance signal (Sec.~\ref{sec:path_following}), and compare it against reactive baselines and \model~operating in planning mode (Sec.~\ref{sec:planning_exp}). Table~\ref{tab:local_comparison} reports Success Rate, Traversal Time, and Interventions on Spot across three scenarios -- Obstructed Goal, Long Distance, and Indoor Corridor -- for three reactive baselines (GNM, ViNT, and NoMaD), VertiFormer, \model's two local path following variants (CFG and Sampling), and \model's planner.

\input{contents/policy_comparison_table}

\textbf{\model~matches reactive baselines despite sparser guidance.} GNM, ViNT, and NoMaD have the advantage of accessing a topological map providing a dense stream of intermediate goal images at 2Hz. Conversely, 
\model's variants receive only sparse, widely spaced waypoints; despite this, \model~matches or exceeds the reactive baselines' success rate and intervention counts across all scenarios tested.

\textbf{CFG and Sampling trade off differently across horizons.} CFG achieves the lowest traversal time and zero interventions on Indoor Corridor, but degrades sharply on Long Distance (138 $\pm$ 7.7s, 2.4 $\pm$ 0.9 interventions): because $\tilde{z}_g$ is computed via continuous interpolation between $z_g$ and $z_\emptyset$ (Sec.~\ref{sec:path_following}), small per-step deviations compound over long horizons, eventually pushing the latent outside the valid VQ manifold and triggering hallucinated dynamics. Sampling avoids this by restricting sampling candidates to the pose codebook, making it a more reliable choice for long-horizon following at some cost in traversal time.

\textbf{VertiFormer remains a strong baseline}, achieving the fastest traversal time on Obstructed Goal (21.0 $\pm$ 1.05s) and matching the zero intervention count on Indoor Corridor, consistent with a deterministic regression head reaching a single hypothesis directly without search or multi-step rollout, trading collision-avoidance and goal-flexibility (Table~\ref{tab:avoidance_cost_ablation}) for speed. Offline evaluations detailed in App.~\ref{app:dataset_eval} exhibit a similar pattern where deterministic baselines edge out generative methods.

\textbf{Planning vs.\ local path following.} \model's planning mode matches or exceeds CFG and Sampling's success rate on Obstructed Goal (90\%) and Indoor Corridor (100\%), at the cost of higher traversal time and variance from searching at every step. On long distance, however, planning succeeds in only 40\% of trials versus 100\% for CFG and Sampling: the added per-step search latency is enough to make the robot react late at the scenario's single corridor fork, an unrecoverable failure mode given the trial's time limit even though the same latency is harmless in the shorter scenarios. This is a latency-budget mismatch rather than a failure of the discrete search itself; see App.~\ref{app:failures} for details.

%% file: contents/planning_comparison_table.tex
\begin{table}[h]
\centering
\caption{Planning Performance. While VertiFormer with MPPI runs quickly, it suffers from sample inefficiency and is blind to obstacles (0\% SR), as is the continuous variant, Hydra-NoVQ. \model's  DLP achieves high success rates while maintaining real-time execution speeds compared to NWM.}
\vspace{1mm}
\label{tab:global_planning}
\setlength{\tabcolsep}{3pt}
\resizebox{\columnwidth}{!}{%
\begin{tabular}{ll ccc cc cccc}
\toprule
\textbf{Model} & \textbf{Params} & \textbf{Search Space} & \textbf{Iterations} & \textbf{Samples} & \textbf{Len.} & \textbf{Plan Time $\downarrow$} & \textbf{Unobs. (SR) $\uparrow$} & \textbf{Obs. (SR) $\uparrow$} & \textbf{Corner Turn (SR)$\uparrow$}\\
\midrule
NWM/B & 200M & Continuous (CEM) & 3 & 18 & 12 & $>$ 500.0s & 0 & 0 & 0 \\
VertiFormer  & 27.11M & Continuous (MPPI) & 1 & 18 & 12 & \textbf{$\sim$ 0.7s} & 40 & 0 & 0\\
Hydra-NoVQ  & 141.60M & Continuous (MPPI) & 1 & 18 & 12 & $\sim$ 1.2s & 20 & 10 & 10\\
\rowcolor{LightBlueBg} \textbf{\model} & 143.29M & Discrete (DLP) & 3 & 18 & 12 & $\sim$ 0.9s &  \textbf{100} & \textbf{80}  & \textbf{80}\\
\bottomrule
\end{tabular}
}
\end{table}

%% file: contents/policy_comparison_table.tex
\begin{table*}[t]
\centering
\caption{Local Planning Evaluation across 10 Physical Trials per Scenario. \model~drastically reduces physical interventions by leveraging its predictive foresight to move around topological traps that ensnare reactive baselines. GNM, ViNT, and NoMaD are conditioned on dense goal-image sequences at 2Hz; \model's local-following variants (CFG, Sampling) are conditioned on sparse, more widely spaced waypoints.}
\label{tab:local_comparison}
\scalebox{1}{
\resizebox{\textwidth}{!}{%
\begin{tabular}{ll ccc cccc}
\toprule
& & \multicolumn{3}{c}{\textbf{Dense Goal Images (2Hz)}} & \multicolumn{4}{c}{\textbf{Sparse Waypoints (turns only)}} \\
\cmidrule(lr){3-5} \cmidrule(lr){6-9}
\textbf{Scenario} & \textbf{Metric} & \textbf{GNM} & \textbf{ViNT} & \textbf{NoMaD} & \textbf{\model~(CFG)} & \textbf{\model~(Sampling)} & \textbf{VertiFormer} & \textbf{\model~(Planning)}\\
\midrule

\multirow{3}{*}{\textbf{Obstructed Goal}}
 & Success Rate (\%) $\uparrow$
 & 90 & 80 & \textbf{100} & 90 & 50 & \textbf{100} & 90 \\
 & Traversal Time (s) $\downarrow$
 & 36.4 $\pm$ 7.9
 & 34.4 $\pm$ 6.8
 & 37.4 $\pm$ 7.7
 & 45.0 $\pm$ 8.0
 & 48.4 $\pm$ 18.5
 & \textbf{21.0} $\pm$ 1.05
 & 37.2 $\pm$ 21.0\\
 & Interventions $\downarrow$
 & 1.1 $\pm$ 0.3
 & 1.2 $\pm$ 0.6
 & 1.1 $\pm$ 0.3
 & 1.0 $\pm$ 0.7
 & \textbf{0.8} $\pm$ 0.8
 & 1.0 $\pm$ 0.5
 & 1.2 $\pm$ 0.6\\
\midrule

\multirow{3}{*}{\textbf{Long Distance}}
 & Success Rate (\%) $\uparrow$
 & 0 & 0 & 0 & \textbf{100} & \textbf{100} & \textbf{100} & 40 \\
 & Traversal Time (s) $\downarrow$
 & - 
 & - 
 & - 
 & 138 $\pm$ 7.7
 & 107 $\pm$ 8.6
 & 109 $\pm$ 7.6 & 119 $\pm$ 3.8 \\
 & Interventions $\downarrow$
 & - 
 & - 
 & - 
 & 2.4 $\pm$ 0.9
 & 1.2 $\pm$ 0.8
 & 1.0 $\pm$ 1.0 & 3.4 $\pm$ 0.6 \\
\midrule

\multirow{3}{*}{\textbf{Indoor Corridor}}
 & Success Rate (\%) $\uparrow$
 & 40 & 90 & 90 & \textbf{100} & \textbf{100} & \textbf{100} & \textbf{100} \\
 & Traversal Time (s) $\downarrow$
 & 59.0 $\pm$ 4.6
 & 48.1 $\pm$ 3.3
 & 65.0 $\pm$ 21.2
 & \textbf{24.4 $\pm$ 2.01}
 & 28.2 $\pm$ 0.79
 & 26.3 $\pm$ 1.6
 & 33.9 $\pm$ 2.0\\
 & Interventions $\downarrow$
 & \textbf{0.0} $\pm$ 0.0
 & \textbf{0.0} $\pm$ 0.0
 & 0.2 $\pm$ 0.4
 & \textbf{0.0} $\pm$ 0.0
 & \textbf{0.0} $\pm$ 0.0
 & \textbf{0.0} $\pm$ 0.0
 & 0.1 $\pm$ 0.3\\
\bottomrule
\end{tabular}
}
}
\end{table*}

%% file: contents/discussions.tex
\section{Limitations and Future Work}\label{sec:limitations}

While \model~demonstrates advancements in latent planning, empirical deployment reveals challenges in DLP. We suggest reading App.~\ref{app:failures} for a further discussion of the limitations.

\textbf{Manifold Rejection Confuses Visual Complexity with Genuine Risk.}
Manifold Rejection ($\mathcal{C}_{vq}$) is a useful but imperfect safety signal: it cannot fully distinguish a genuinely dangerous occlusion from a visually complex but entirely safe scene. As shown empirically in Fig.~\ref{fig:vq_energy_spike}, dense, high-frequency textures such as foliage and grass produce quantization error nearly as large as a true physical obstacle, since both are difficult for the visual codebook to reconstruct precisely, one because it is genuinely unfamiliar and dangerous, the other simply because it is visually intricate. We refer the readers to App.~\ref{sec:manifold_rejection_eval} for more details. 

\textbf{Representational Capacity Trades Expressivity for Latency.}
\model's zero-shot generalization is intrinsically tied to the vocabulary size of its VQ codebooks. A larger codebook would capture a wider distribution of visual and kinodynamic behaviors, improving generalization to novel environments, but at a direct computational cost: retrieving the nearest codebook entry during search requires a distance computation whose cost scales with vocabulary size, contradicting the same real-time latency budget that motivated discretization in the first place (Sec.~\ref{sec:planning_exp}). \model~therefore faces a structural trade-off between behavioral expressivity and search latency that a fixed-size codebook cannot resolve on its own; adaptive or hierarchical vocabularies are a natural direction for future work.

\section{Conclusion}\label{sec:conclusion}
We presented \model, a World Action Model that closes the representation misalignment between generative world models and their planners by moving the planner inside the model's own discrete latent manifold. Across two physical robots, this yields a $\sim$500$\times$ reduction in planning latency over NWM while achieving higher success rates in obstacle-rich environments than every baseline we tested, including \model-NoVQ, isolating discretization itself as the source of this gain rather than model capacity or Flow Matching. These results suggest that discretizing a robot's intent is a promising paradigm for scaling world models to real-time physical deployment. As our limitations discuss, realizing this promise more broadly will require representations expressive enough to avoid the failure modes a fixed-size codebook currently introduces, particularly for dynamic agents and finer-grained obstacle discrimination.

%% file: contents/acknowledgment.tex
\section*{Acknowledgment}
This work is a collaboration between the RobotiXX Lab, the VCAI Lab, and the AIR Lab. Research at the RobotiXX Laboratory is supported by the National Science Foundation (NSF, 2350352), the Army Research Office (ARO, W911NF2320004, W911NF2520011), the Army Ground Vehicle Systems Center (GVSC), Google DeepMind (GDM), Microsoft Research (MSR), Clearpath Robotics, FrodoBots Lab, Raytheon Technologies (RTX), Tangenta, 4-VA, the Mason Innovation Exchange (MIX), and Walmart. Research by the VCAI Lab was conducted under the RobOdin project, funded by the European Union through its Interreg Germany-Denmark funding program, with technical support provided by the Embodied AI Center at Kiel University. Work at the AIR Lab was supported by the Institute of Information \& Communications Technology Planning \& Evaluation (IITP) and Information Technology Research Center (ITRC) grant, funded by the Korean government (Ministry of Science and ICT) (IITP-2026-RS-2020-II201460).

%% file: contents/appendix.tex
\newpage
\appendix
\section*{Appendix}

Table~\ref{tab:comparison} summarizes how \model~relates to prior navigation baselines across five dimensions -- observation input, goal input, conditioning input, output modalities, and navigation policy type -- extending the architectural comparison of Fig.~\ref{fig:arch_baselines}.

\input{contents/comparison_table}

\section{Problem Formulation}\label{app:problem_formulation}

\textbf{Reactive Policy Formulation:}
Reactive foundation models (\eg, ViNT~\citep{shah2023vint}, GNM~\citep{shah2023gnm}, and NoMaD~\citep{sridhar2023nomad}) frame navigation as a direct sequence-to-sequence mapping problem. Given the observation history $\mathcal{H}_t$ and a goal representation $\mathcal{G}$ (typically a target image), the model $f_\pi$ directly outputs a sequence of immediate future actions or spatial waypoints of length $H$:
\begin{equation}
    \hat{a}_{t:t+H-1} = f_\pi(\mathcal{H}_t, \mathcal{G}).
    \label{eq:reactive_policy}
\end{equation}
While computationally lightweight and effective for local trajectory following, reactive policies lack foresight. They cannot predict the future visual state or consequences of their actions, making them susceptible to local minima and rendering them incapable of planning.

\textbf{Visual World Model Formulation (Decoupled Planning):}
To overcome the myopia of reactive policies, WMs (\eg, NWM~\citep{bar2025navigation}) frame navigation as a forward conditional dynamics task. Given the history $\mathcal{H}_t$ and a stochastically sampled sequence of future actions $a_{t:t+H-1}$ from a sampling-based planner, the model $P_W$ generates the expected future visual states, $\hat{I}$:
\begin{equation}
    \hat{I}_{t+1:t+H} \sim P_W(\cdot \mid \mathcal{H}_t, a_{t:t+H-1}).
    \label{eq:wm_dynamics}
\end{equation}
This formulation requires Model Predictive Control (MPC) where the planner and WM operate on two separate manifolds. Because it predicts visual states conditioned on continuous actions, it relies on perceptual metrics to evaluate goal reaching, which is brittle because it does not consider the distance~\citep{ai2025review, tian2023control, bear2021physion}. 

\section{Extended Results}
\label{app:extended_results}

\subsection{Single-Step Image Quality}
\label{app:image_quality}

Following NWM, we evaluate single-step visual predictive fidelity on a held-out subset of the RECON dataset. We benchmark \model~against NWM/B and VertiFormer. As shown in Table~\ref{tab:image_eval}, \model~achieves competitive performance (DreamSim, LPIPS) despite a smaller parameter footprint and encoding multiple modalities with only 50 NFEs compared to NWM's 250 NFEs. While VertiFormer achieves the highest PSNR, this metric is notoriously misaligned with dynamic perception: deterministic models minimize MSE by averaging all possible futures, resulting in severe ``ghosting'' artifacts and blurry textures~\citep{luc2017predict, luc2018predict}. PSNR strictly penalizes pixel-level noise but is blind to structural semantics, causing a higher score for VertiFormer. By leveraging Flow Matching over a discrete bottleneck, \model~avoids regression-to-the-mean, preserving sharp, semantically valid structures and yielding better perceptual scores (LPIPS, DreamSim).

\begin{table}[h]
\centering
\caption{Offline single-step image quality on a held-out subset of the RECON dataset. \model~achieves competitive accuracy with fewer parameters and NFEs.}
\label{tab:image_eval}
\begin{tabular}{ll ccc}
\toprule
\textbf{Model} & \textbf{Params} & DreamSim $\downarrow$ & LPIPS $\downarrow$ & PSNR $\uparrow$  \\
\midrule
NWM/B & 200M & 0.2480 $\pm$ 0.1265 & 0.5168 $\pm$ 0.1147 & 10.69 \\
VertiFormer & 27.11M & 0.3738 $\pm$ 0.0560 & 0.5417 $\pm$ 0.0757 & \textbf{18.29}\\
\rowcolor{LightBlueBg} \textbf{\model} & 143.29M & \textbf{0.1103} $\pm$ 0.0345 & \textbf{0.2809} $\pm$ 0.0679 & 16.20\\
\bottomrule
\end{tabular}
\end{table}

\input{contents/dataset_comparison_table}

\subsection{Dataset Evaluation}\label{app:dataset_eval}
Table~\ref{tab:offline_eval} reports offline trajectory forecasting accuracy (ADE/FDE) on a held-out subset across three navigation datasets (SCAND, TartanDrive, and SACSoN); VertiFormer, a deterministic baseline, edges out generative methods on these metrics. However, as reported in Table~\ref{tab:local_comparison}, offline performance does not reliably indicate real-world success. Offline evaluations inherently assume fixed trajectory lengths and guaranteed short-term goal reachability. Conversely, physical execution suffers from compounding centimeter-scale errors that progressively deviate the robot from its intended trajectory.

\subsection{Long-Horizon Qualitative Comparison}
\label{app:qualitative}

\begin{figure}[h]
    \centering
    \includegraphics[width=\linewidth]{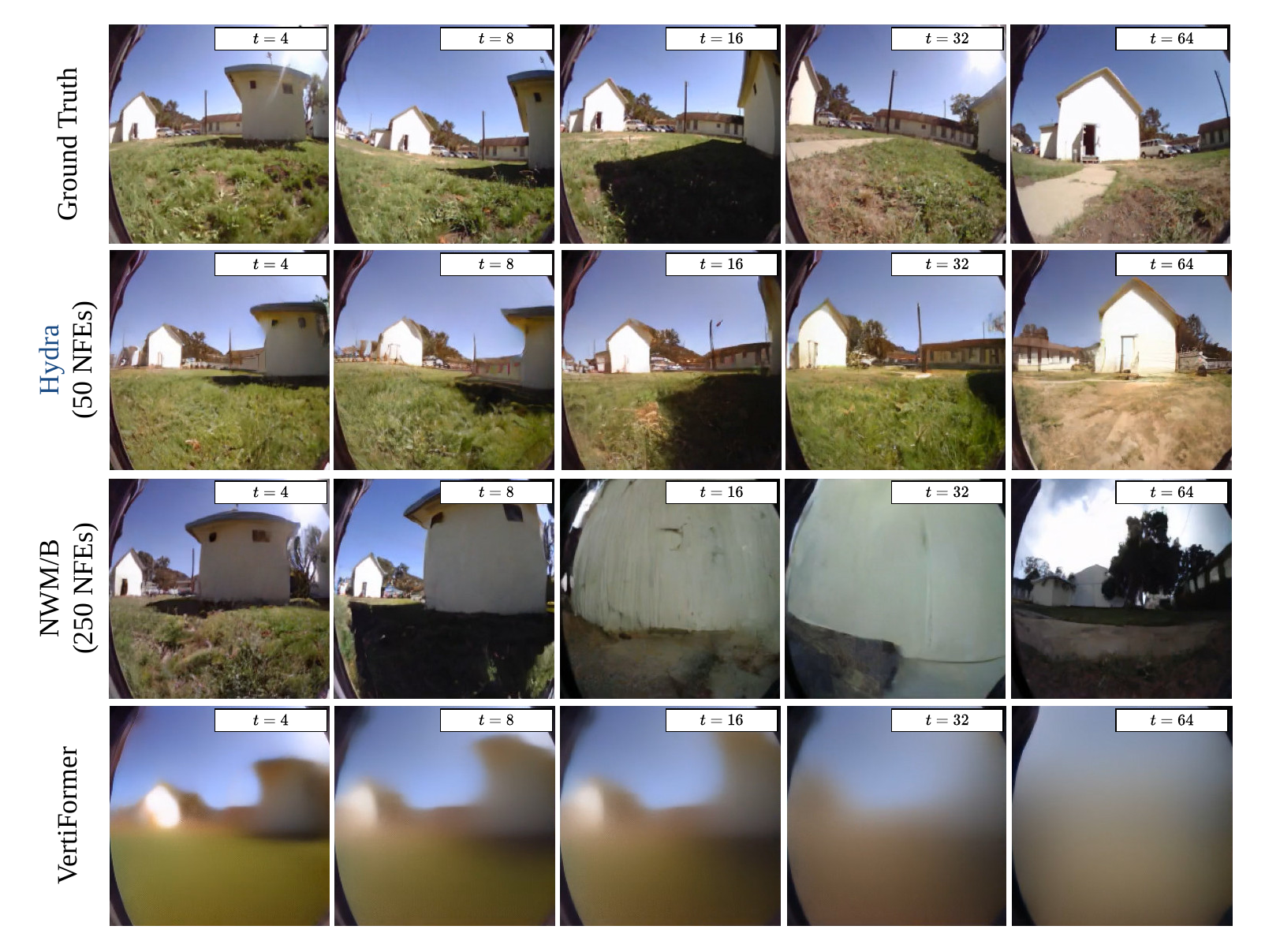}
    \caption{Comparison of a 16-second video generated from a held-out subset of RECON. While NWM's continuous latents drift from the commanded actions over time, \model's discrete manifold maintains action adherence over time.}
    \label{fig:qualitative}
\end{figure}

Figure~\ref{fig:qualitative} shows the qualitative counterpart to the FVD comparison in Table~\ref{tab:fvd} (Sec.~\ref{sec:video_comparison}). NWM initially renders high-fidelity frames, but its continuous latent space suffers from compounding errors; after approximately 8 seconds, the visual context rapidly degrades and decouples from the commanded actions (hallucination). \model's discrete intent manifold adheres accurately to the physical commands over the full 16-second horizon.

\subsection{KPC Ablations}
\label{sec:ablations}

We ablate the individual terms of $\mathcal{J}_{\text{KPC}}$ (\eqref{eq:kpc}) to isolate which components drive \model's obstacle-avoidance behavior. All results are averaged over 10 physical trials per configuration.

\begin{wraptable}{r}{0.5\textwidth}
  \centering
  \caption{Effect of the image-goal cost term ($\lambda_{\text{img}}$) conditioned on goal visibility. The same term that improves performance when the goal is visible degrades it when the goal is not, motivating conditioning $\mathcal{C}_{\text{img}}$ on goal visibility rather than applying it unconditionally.}
  \label{tab:goal_visibility_ablation}
  \resizebox{\linewidth}{!}{%
  \begin{tabular}{lcc}
  \toprule
  \textbf{Scenario} & \textbf{SR (\%) $\uparrow$} & \textbf{Time (s) $\downarrow$} \\
  \midrule
  Frontal, $\lambda_{\text{img}} = 1$        & 80 & 24.7 $\pm$ 5.8\\
  Blind corner, $\lambda_{\text{img}} = 1$  & 20 & 30.6 $\pm$ 12.1 \\
  Blind corner, $\lambda_{\text{img}} = 0$ & 70 & 34.5 $\pm$ 8.9   \\ 
  \bottomrule
  \end{tabular}%
  }
\end{wraptable}

\textbf{Image-goal term.} We first ask whether $\mathcal{C}_{\text{img}}$ should be active unconditionally. Table~\ref{tab:goal_visibility_ablation} compares two scenarios that differ only in whether the goal is visible from the ego camera: frontal obstacle avoidance, where the goal remains at least partially in view, and a blind corner turn, where the goal is entirely outside the camera view until after the turn. With $\mathcal{C}_{\text{img}}$ active in both cases, success rate drops from 80\% when the goal is visible to 20\% when it is not. Disabling the goal term when the goal is not visible ($\lambda_{\text{img}} = 0 $) improves the success rate.
Without a true visual anchor for the goal, the term can be minimized by any visually plausible candidate rather than the geometrically correct one, actively misleading the planner rather than aiding it. Based on this result, in the subsequent experiment (Table~\ref{tab:avoidance_cost_ablation}), $\mathcal{C}_{\text{img}}$ is disabled.

\textbf{The remaining avoidance terms are individually necessary and embodiment-agnostic.} With $\mathcal{C}_{\text{img}}$ handled as above, Table~\ref{tab:avoidance_cost_ablation} ablates the three terms, $\mathcal{C}_{\text{prior}}$, $\mathcal{C}_{\text{vq}}$, and $\mathcal{C}_{\text{ent}}$, individually, on both Spot and Jackal. Removing any single term increases collisions relative to the full cost on both platforms, with the effect even more pronounced on Jackal (0.20 rising as high as 0.80) than on Spot (0.20 rising to 0.40--0.70). That the same qualitative degradation appears on two robots with different kinodynamics indicates each term contributes a distinct, non-redundant safety signal.

Notably, $\mathcal{C}_{\text{prior}}$'s contribution follows from the unconditional expert prior being fit only to collision-free human demonstrations: a trajectory that intersects an obstacle is itself far outside this distribution. $\mathcal{C}_{\text{prior}}$ therefore penalizes collisions as a consequence of penalizing implausibility under the expert distribution, complementing $\mathcal{C}_{\text{vq}}$'s explicit OOD rejection in the visual codebook rather than duplicating it.

\begin{table}[t]
\centering
\caption{Ablation of obstacle-avoidance cost terms ($\lambda_{\text{prior}}$, $\lambda_{\text{vq}}$, $\lambda_{\text{entropy}}$), across two robot embodiments: a blind corner turn (Spot) and frontal obstacle avoidance (Jackal). Each avoidance term degrades performance when removed on \emph{both} platforms, indicating the mechanism is embodiment-agnostic.}
\label{tab:avoidance_cost_ablation}
\begin{tabular}{l cc cc}
\toprule
& \multicolumn{2}{c}{\textbf{Spot}} & \multicolumn{2}{c}{\textbf{Jackal}} \\
\cmidrule(lr){2-3} \cmidrule(lr){4-5}
\textbf{Configuration} & \textbf{Time (s) $\downarrow$} & \textbf{Collisions $\downarrow$} & \textbf{Time (s) $\downarrow$} & \textbf{Collisions $\downarrow$} \\
\midrule
Original (avoidance costs + $\mathcal{C}_{\text{geo}}$) & 30.6 $\pm$ 12.1 & \textbf{0.20 $\pm$ 0.42} & 23.8 $\pm$ 0.39 & \textbf{0.20 $\pm$ 0.13}\\
$\lambda_{\text{prior}} = 0$                              & 33.9 $\pm$ 13.0 & 0.70 $\pm$ 0.48        & 21.2 $\pm$ 0.26 & 0.70 $\pm$ 0.59 \\
$\lambda_{\text{vq}} = 0$                                 & 30.3 $\pm$ 10.0 & 0.50 $\pm$ 0.53        & 20.9 $\pm$ 0.34 & 0.60 $\pm$ 0.28 \\
$\lambda_{\text{entropy}} = 0$                            & 36.2 $\pm$ 18.9 & 0.40 $\pm$ 0.52        & 26.6 $\pm$ 0.45 & 0.80 $\pm$ 0.61 \\
\bottomrule
\end{tabular}
\end{table}

\subsection{Analysis of the Kinodynamic Codebook}
\label{sec:codebook_analysis}

\begin{figure}[h]
\centering
    \includegraphics[width=0.7\linewidth]{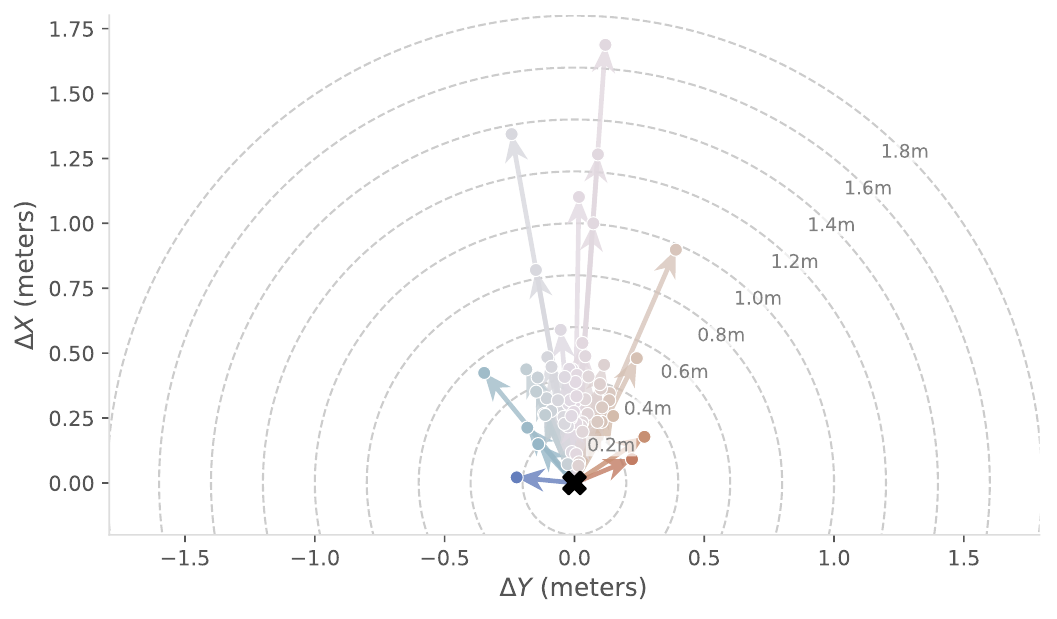}
  \caption{1-step spatial coverage of the learned pose codebook $\mathcal{V}_p$. Unlike normalized baselines, the discrete manifold explicitly captures the absolute physical scaling of different robotic platforms, ranging from low-speed indoor maneuvers (dense central clusters) to high-speed off-road dynamics (sparse outer bounds).}
  \label{fig:codebook_coverage}
\end{figure}

While a continuous action space preserves infinite-resolution control, which is highly beneficial for fine-grained maneuvers if the transition model is perfect, it is sample-inefficient. \model~mitigates the loss of fine-grained continuous resolution by using continuous Flow-Matching during final execution. These primitive discretized actions can be considered the mean of the trajectory decoded by the FM head. We analyze the 1-step spatial geometry of the learned discrete pose codebook, $\mathcal{V}_p$. As shown in Figure~\ref{fig:codebook_coverage}, the codebook captures a diverse, unnormalized distribution of spatial displacements ($\Delta X, \Delta Y$).

On the other hand, a predefined or uniformly spaced grid of actions is fixed independently of the world model. Because such a grid is never trained jointly with the model's own representation, its entries have no defined relationship to the model's embedding space at all, a hand-specified ``turn left'' primitive is not a point the encoder ever produced, so there is no meaningful notion of its distance to anything the model has learned. By contrast, because \model's codebook is learned end-to-end alongside perception and dynamics, every entry is, by construction, a point the model itself produced and can situate relative to its own manifold. The distance metric between codebook entries is therefore meaningful only because the codebook and the world model share one coupled representation. This is the same measure of safety that the KPC cost uses to evaluate candidate trajectories in the latent space without decoding during the planning phase.

\textbf{The Flaw of Action Normalization.}
Generalist navigation models (\eg, GNM, ViNT, NoMaD, and NWM) universally normalize actions by the maximum speed of the specific robot platform to achieve dataset-agnostic training. However, this normalization destroys the absolute scale of the physical world; a maximum-throttle command for a slow indoor crawler becomes mathematically indistinguishable from a maximum-throttle command for a high-speed off-road ATV. Consequently, the generative model becomes blind to the physical constraints of the environment.

\textbf{Emergent Physical Grounding.}
By using unnormalized actions, \model's codebook naturally segregates physical domains. Low-speed datasets populate the dense inner clusters of $\mathcal{V}_p$ ($<0.5m$), while high-speed datasets form the sparse outer bounds ($>1.0m$). Because the observation history $\mathcal{H}_t$ is contextually fused with these absolute velocities, \model~learns a strict correlation between visual textures and physical speed limits.

This unnormalized grounding explains an emergent behavior in \model: \textit{Contextual Manifold Shift}. If the model is conditioned on an indoor visual history (implying a slow-moving platform like a Clearpath Jackal) but is suddenly forced to execute a high-speed action intent (e.g., $1.5$ m/step), the model will abruptly ``hallucinate'' a transition to an off-road TartanDrive environment. Rather than a standard generative failure, this is a resolution of conflicting modalities: the generative solver recognizes that the commanded velocity has zero probability density within the indoor contextual manifold, and shifts the predicted visual flow to the only environmental sub-manifold where that velocity is physically valid.

By bounding trajectory searches to this physically grounded codebook (\eqref{eq:kpc}), \model~prevents the execution of these OOD speed-context mismatches, ensuring safety in a way normalized models cannot. Furthermore, knowing which codes correspond to high-speed or turning, we can effectively change the robot's behavior during deployment by masking specific codes. For example, masking high-speed codes prevents the robot from driving fast, or masking right turns prevents the robot from turning right.

\subsection{Visualizing Latent Search: RViz Branch Trees}
\label{app:rviz_branching}

To make Discrete Latent Planning's search process (Algorithm~\ref{alg:energy_intent_search}, Appendix~\ref{app:stochastic_planning}) inspectable, we visualize it in RViz during physical deployment.

\begin{figure}[t]
    \centering
    \includegraphics[width=0.9\linewidth]{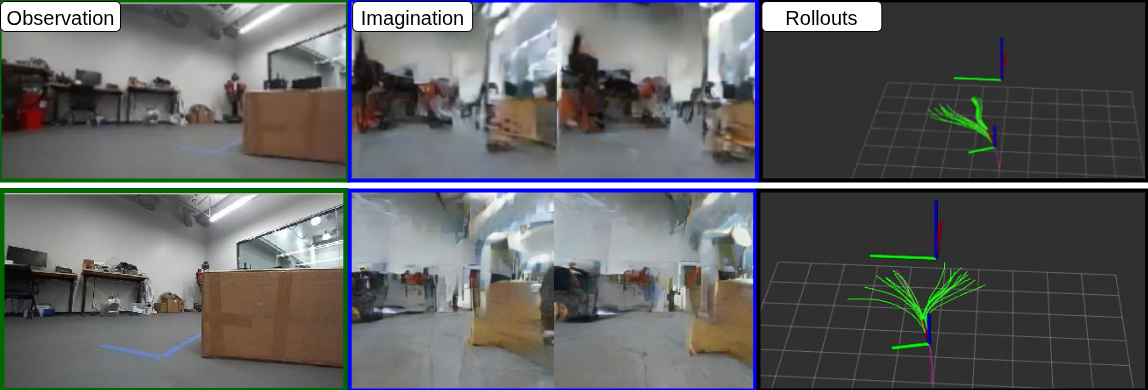}
    \caption{Live RViz visualization of \model's Discrete Latent Planning search on the physical robot in a never-before-seen environment. \textbf{Left:} the robot's current ego-centric observation. \textbf{Middle:} a subset of \model's predicted future frames. \textbf{Right:} the corresponding candidate trajectories (\textcolor{green}{green}) rendered as a branching tree in the robot's local frame; each branch is one topological maneuver considered before $\mathcal{J}_{\text{KPC}}$ selects the lowest-cost (\textcolor{purple}{purple}) candidate for execution.}
    \label{fig:rviz_branching}
\end{figure}

Figure~\ref{fig:rviz_branching} shows a representative search: the search tree fans out from the robot's current pose into a small number of distinct trajectories reflecting the discrete codebook's role in constraining candidates to a small vocabulary of plausible maneuvers (Sec.~\ref{sec:codebook_analysis}) rather than an unconstrained continuous spread.

\subsection{Latent Collision Detection}
\label{sec:manifold_rejection_eval}

\begin{figure}[t]
    \centering
    \includegraphics[width=0.9\linewidth]{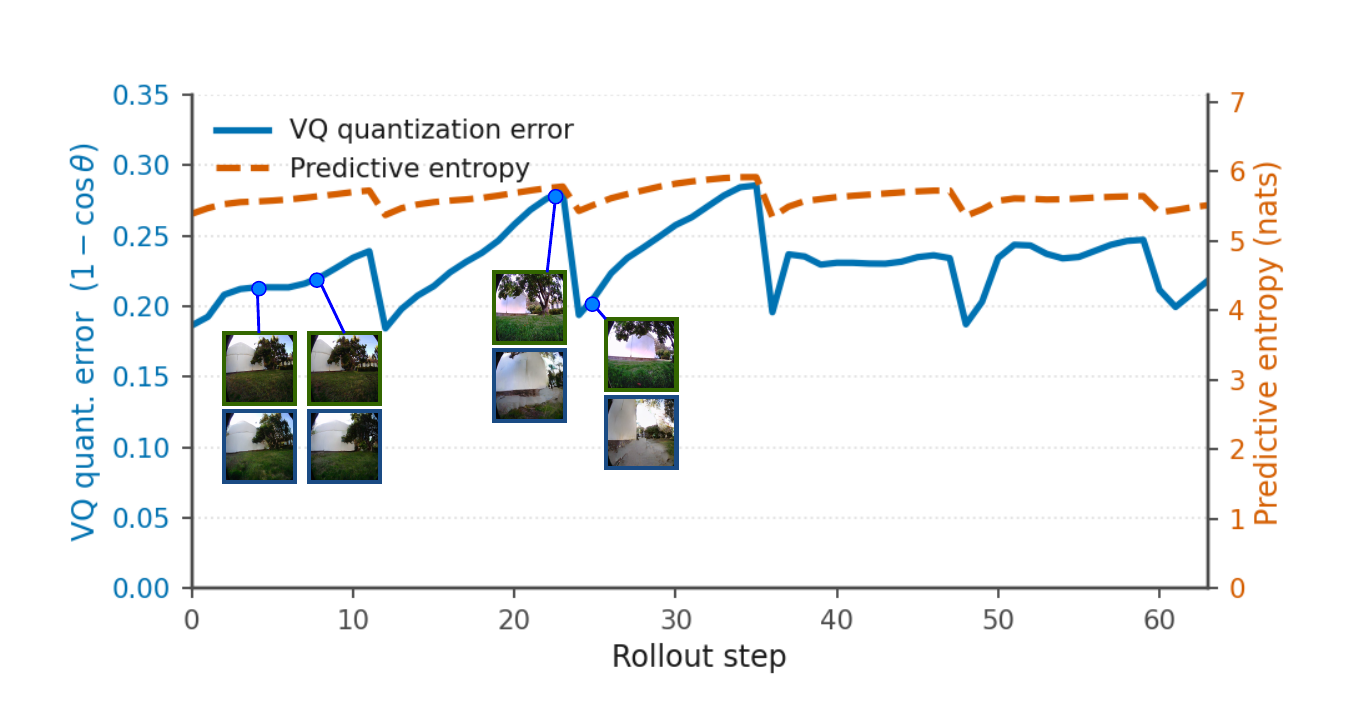}
    \caption{\textbf{Manifold Rejection and Predictive Entropy During Impending Collision.} \textcolor{green}{Top}/\textcolor{HydraBlue}{Bottom} insets: Comparison of \textcolor{green}{Ground Truth} vs. \model's \textcolor{HydraBlue}{imagination} at points along a rollout commanded to drive straight toward a physical occlusion. Both the spatially-averaged VQ quantization error ($\mathcal{C}_{vq}$, solid blue, left axis) and the predictive entropy of the visual token distribution ($\mathcal{C}_{ent}$, dashed orange, right axis) rise together as the commanded trajectory approaches the occlusion, and drop together immediately after the model hallucinates a navigable resolution, two independently computed signals converging on the same collision-relevant moments. The smaller early fluctuation (first two insets) is a false positive triggered by visually complex but traversable foliage, not a genuine obstacle.}
    \label{fig:vq_energy_spike}
\end{figure}

To empirically examine Manifold Rejection ($\mathcal{C}_{vq}$) and Predictive Entropy ($\mathcal{C}_{ent}$) as complementary latent signals correlated with collision risk, we visualize both signals jointly for \model~when forced into a geometric contradiction: a trajectory commanded to drive straight into an obstacle (Fig.~\ref{fig:vq_energy_spike}).

We compute the VQ quantization error of the visual tokens and the Shannon entropy of their distribution at each rollout step. As the commanded trajectory approaches the obstacle, the continuous intent vector drifts toward the undefined space between valid discrete codebook anchors, and both signals rise in tandem: quantization error rises because the predicted embedding has no close codebook match, and entropy rises because the model becomes less confident about which future is plausible. When the generative head is ultimately forced to resolve this OOD state, a \textit{Contextual Manifold Shift}, by hallucinating a navigable pavement in place of the occlusion, both signals drop together immediately afterward, because the fabricated resolution is once again close to something the model has seen and is confident about.

The signal is not perfectly discriminating, however. The smaller fluctuation in $\mathcal{C}_{vq}$ visible earlier in the rollout is a false positive: the model registers elevated quantization error while approaching visually complex but entirely traversable foliage, rather than a genuine obstacle. This introduces a limitation: dense, textured content such as leaves and grass can locally resemble the reconstruction difficulty of a true occlusion, even though no collision risk is present. \model's OOD signal is therefore a strong, but not infallible, indicator of collision risk. The reverse failure is also possible: homogeneous obstacles such as flat walls are comparatively easy to reconstruct and may therefore be under-flagged despite being genuine hazards. Because the present signal cannot separate \emph{why} a state is hard to reconstruct, occasional false positives and potentially false negatives are unavoidable with the current architecture. Addressing this will likely require a representation that distinguishes structural novelty that should be avoided from complex visual texture that is safe to traverse, rather than collapsing both into a single quantization-error signal.

This joint behavior is nonetheless the crux of \model's DLP. Because $\mathcal{C}_{vq}$ and $\mathcal{C}_{ent}$ are computed independently, one from quantization geometry, one from predictive uncertainty, but rise and fall together at exactly the moments that matter, heavily weighting both in the KPC (\eqref{eq:kpc}) lets the optimizer reject a candidate trajectory \textit{before} the generative head is forced to hallucinate, using two agreeing, pixel-free signals rather than one. Occasional false positives aside, this remains substantially more reliable than the alternative of not evaluating safety at all until pixels are rendered: the entire mechanism, ranking, and rejection, happens in latent space, extracting a safety constraint directly from the quantization and uncertainty of the visual world model without ever decoding to pixels.

\subsection{Failure Analysis}
\label{app:failures}

\begin{figure}[h]
    \centering
    \includegraphics[width=0.8\linewidth]{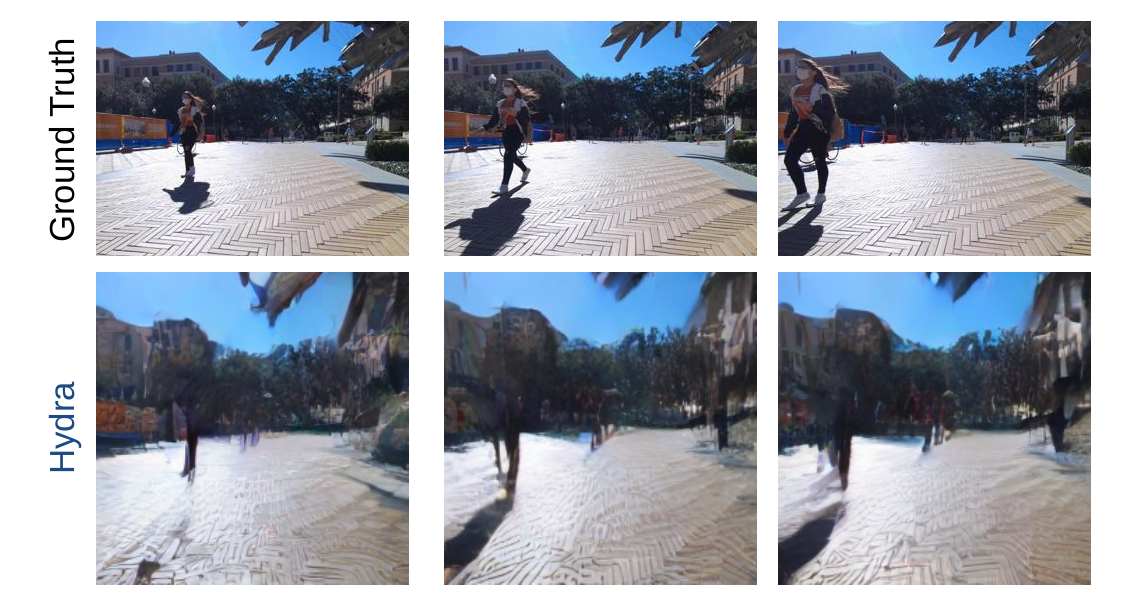}
    \caption{Vanishing humans in future predictions on SCAND. Top: ground-truth future frames, in which a pedestrian remains visible throughout. Bottom: \model's predicted frames for the same sequence, in which the pedestrian is smoothed into the background.}
    \label{fig:scand_failure}
\end{figure}

\paragraph{Failure mode: missed pedestrians.}
Figure~\ref{fig:scand_failure} shows a representative failure mode on SCAND's social navigation scenarios: pedestrians present in the ground-truth future are absent from \model's prediction. We attribute this to the limited cardinality of the visual codebook ($|\mathcal{V}_i| = 2048$), which acts as a stochastic low-pass filter: humans occupy a small subset of pixels relative to the static scene, and are discarded as high-frequency detail during quantization, the same limitation noted in our main-text discussion of limitations. This issue exists in NWM as well, but at a lower frequency. Incorporating human-centric object detection or increasing visual codebook capacity are natural directions to address this in future work.
 
\paragraph{Failure mode: goal-blind seed initialization.}
\begin{figure}[h]
    \centering
    \includegraphics[width=0.8\linewidth]{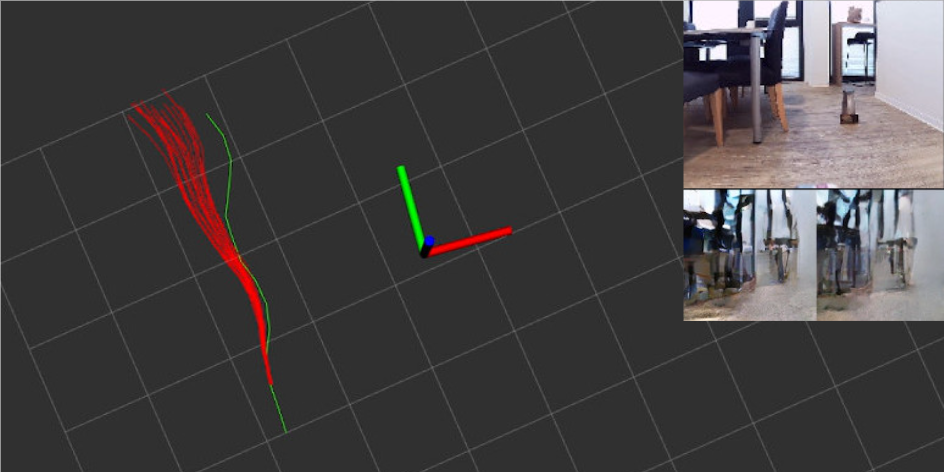}
    \caption{Goal-blind seed initialization causes a missed turn. The green trajectory is the path selected by the DLP cost among the sampled candidates at convergence; the seed distribution that generated the candidates was conditioned only on local scene context, not on the goal, delaying goal-aware correction until it is too late to complete the turn.}
    \label{fig:seed_failure}
\end{figure}
 
In Fig.~\ref{fig:seed_failure}, the green trajectory shows the path selected by the DLP cost among the sampled candidates. This failure arises from seed conditioning: the seed distribution is generated without knowledge of the goal, so goal information only enters the optimization indirectly, through the Euclidean-distance term of the KPC cost, which is evaluated \emph{after} each iteration's candidates are scored and only biases the seed population for the \emph{next} iteration. With $N=3$ iterations, set by the real-time planning budget, this leaves little runway for goal information to propagate before the robot must act. When a required turn falls close to the start of the planning horizon, the robot can physically pass the turning point before the distance signal has had enough iterations to steer the search toward it.
 
\paragraph{Design tradeoff: goal-conditioned vs. unbiased seeding.}
This failure mode reflects a more general tension between convergence speed and search diversity. We conditioned the seed on the goal direction. This resolved the near-turn failure directly, since the search no longer needed multiple iterations to discover the goal direction. However, it introduced a different problem: with all seeds already pointed at the goal, the sampled candidates clustered close together in trajectory space, reducing the diversity of the search and limiting exploration of the KPC cost landscape. We chose unbiased seed conditioning as the more conservative choice, accepting the occasional near-turn failure mode described above in exchange for preserving exploration diversity across the sample population. We leave resolving this tradeoff, for instance via coarser goal-direction (rather than full goal-pose) conditioning, or an explicit diversity term applied alongside goal-conditioning, to future work.

\section{Learning Perspective on \model}
\label{sec:learning_perspective}

As shown in Figure~\ref{fig:teaser}, \model~moves planning from an external module that queries a separately trained world model to a planner that lives inside the world model. We discuss the different parts of \model~from a learning perspective that aided this transition.

\textbf{Joint Distribution vs. Marginal Alignment.} From a representation learning perspective~\citep{buchanan2025learning}, traditional late-fusion architectures optimize isolated marginal distributions and attempt to align them post-hoc~\citep{liu2023visual} to prevent the dominant modality's gradients from overcoming the other~\citep{liu2026tuna2}, often requiring complex contrastive losses. TUNA-2~\citep{liu2026tuna2} showed that this causes representation misalignment, however, given a large amount of data, the model can learn to align different modalities end-to-end. VertiFormer~\citep{nazeri2025vertiformer}~showed that, by employing early fusion, a model is able to approximate the joint multimodal distribution $p(I, p, a)$ on limited data. This imposes an inductive bias on the Transformer to learn the underlying physical correlations (\eg, forward kinematics and inverse dynamics) from the unified token sequence. Furthermore, the curriculum-based random infilling acts as a generalized Denoising Autoencoder objective~\citep{vincent2008extracting}, aiding the network in learning representations that smoothly interpolate across missing sensor modalities.

\textbf{Information Bottleneck and Posterior Collapse.} The necessity of the discrete VQ bottleneck can be analyzed through the lens of the Information Bottleneck principle \citep{tishby2015deep}. Unbounded continuous latent spaces in autoregressive or generative sequence models are notoriously prone to posterior collapse or high-frequency hallucination, as the model can perfectly memorize noisy inputs without extracting underlying causal structure. By forcing the continuous intent through a fixed-size VQ codebook, we bound the mutual information between the observation history and the generated future. This information constraint forces the network to discard accumulated low-level motor noise between prediction steps and compress interactive dynamics into a compact manifold of semantically meaningful kinodynamic primitives.

With this objective in mind, because these codebook embeddings are initialized randomly, passing them directly into an uncalibrated continuous differential equation solver causes rapid divergence. Therefore, we introduce a deterministic linear anchoring that ensures the codebook manifold remains stable~\citep{jing2021understanding} while the downstream generative ODE learns to map this intent to continuous trajectories. Before the flow-matching stage, we apply an MSE loss to align the transformer output and the embedding of the ground truth modality. This is similar to pretraining the model on next-frame prediction, then fine-tuning it to predict high-quality images using flow matching; however, we do both stages at the same time, mimicking the REPA~\citep{yu2024representation} regularization term.

A deep, non-linear projection would possess sufficient capacity to untangle a disorganized latent space~\citep{chen2020simple}, inadvertently permitting the VQ codebook to remain uninformative and unstructured, whereas a linear projection forces implicit subspace whitening on the discrete codebook~\citep{chaudhry2026geometry, gupta2022understanding}. Furthermore, a linear projection mathematically compels the discrete embeddings to construct a highly structured, linearly separable manifold early in the training process. This constrained `tight grip' ensures that the conditioning signal provided to the subsequent Flow Matching module acts as a robust, geometrically meaningful prior. We empirically validate the necessity of this strict linear information bottleneck in Appendix \ref{sec:ab:utilization}, demonstrating that it significantly improves codebook utilization and training convergence.

\textbf{Optimal Transport and ODE Stability.} While recent generative policies heavily rely on score-based Diffusion Models \citep{chi2025diffusion}, diffusion is formulated as a Stochastic Differential Equation (SDE) that suffers from highly curved sampling trajectories. \model~utilizes Rectified Flow Matching~\citep{liu2022flow}, which frames the generative process as an Ordinary Differential Equation (ODE). From an optimization standpoint, mapping a Gaussian prior directly to the target distribution via a linear velocity field $u(\psi_s) = \psi_1 - \psi_0$ implicitly encourages Optimal Transport (OT) \citep{lipman2022flow}. This OT property results in straighter, deterministic probability paths during inference. For robotics, this is a critical theoretical advantage: straighter ODE paths allow the model to be integrated with significantly fewer Number of Function Evaluations (NFEs) without discretization error, enabling the real-time inference speeds required for physical deployment. This is in contrast with NWM, which formulates the generations as an SDE.

\textbf{The Perceptual Fallacy: Limitations of LPIPS.}
Prior image-goal navigation frameworks utilizing world models often rely on visual similarity metrics to evaluate predicted trajectories. NWM, for instance, utilizes Learned Perceptual Image Patch Similarity (LPIPS) to calculate the distance between the predicted visual future and the goal image.

\begin{figure}[!h]
    \centering
    \subfloat[\centering Goal Image]{{\includegraphics[width=4cm, height=3cm]{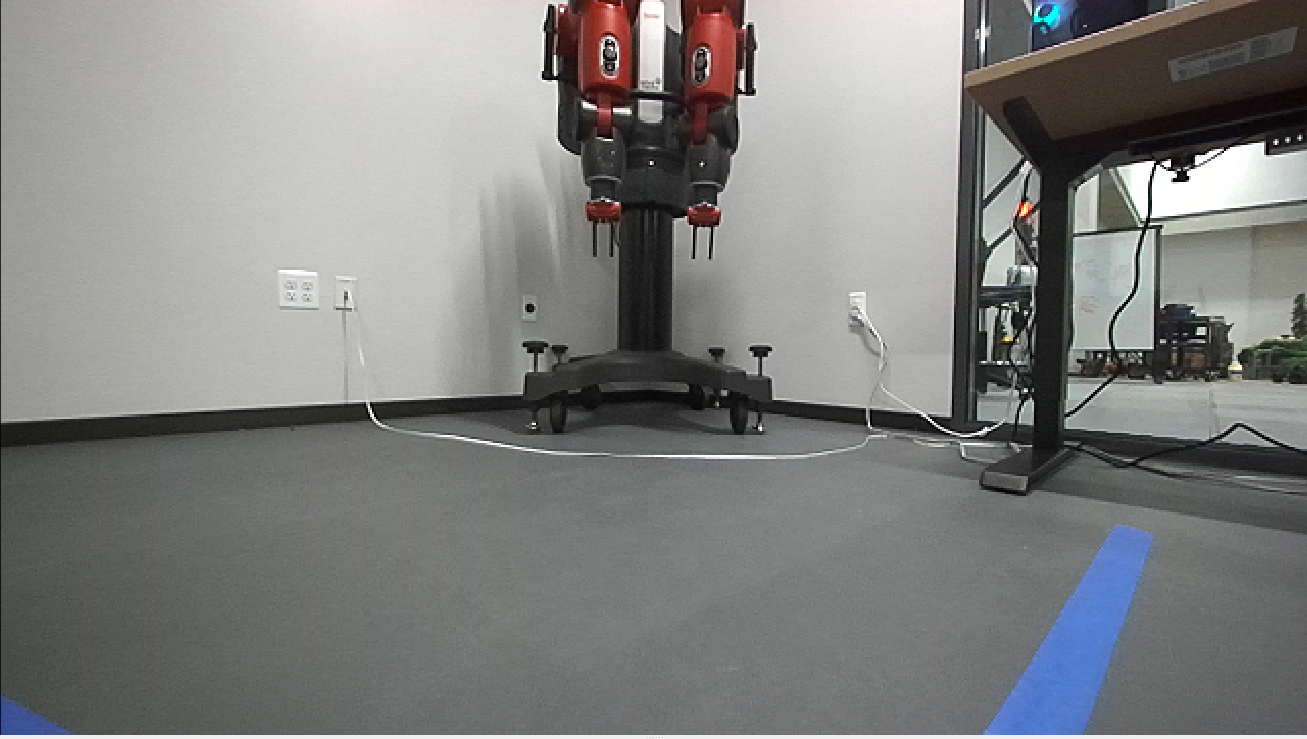} }}%
    \qquad
    \subfloat[\centering LPIPS Loss: 0.50]{{\includegraphics[width=4cm, height=3cm]{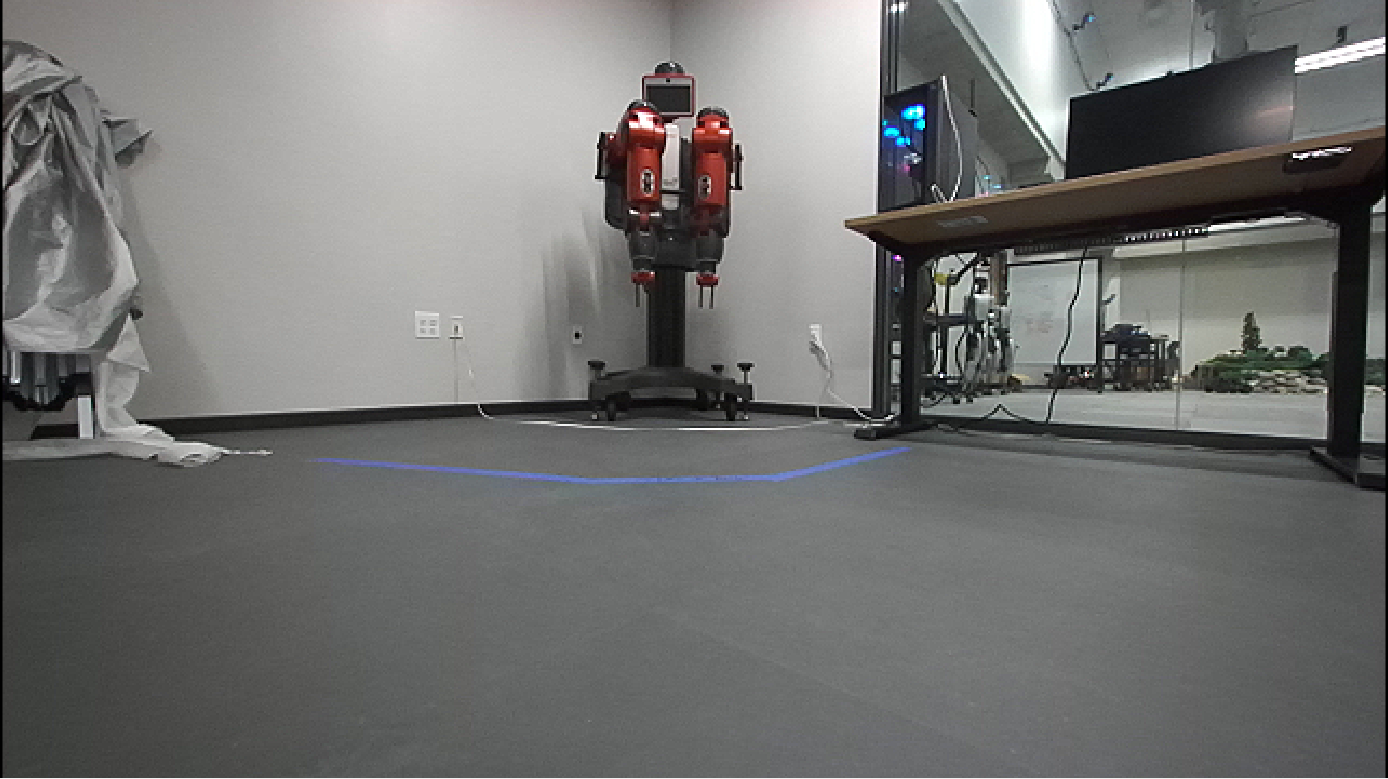} }}
    \qquad
    \subfloat[\centering LPIPS Loss: 0.52]{{\includegraphics[width=4cm, height=3cm]{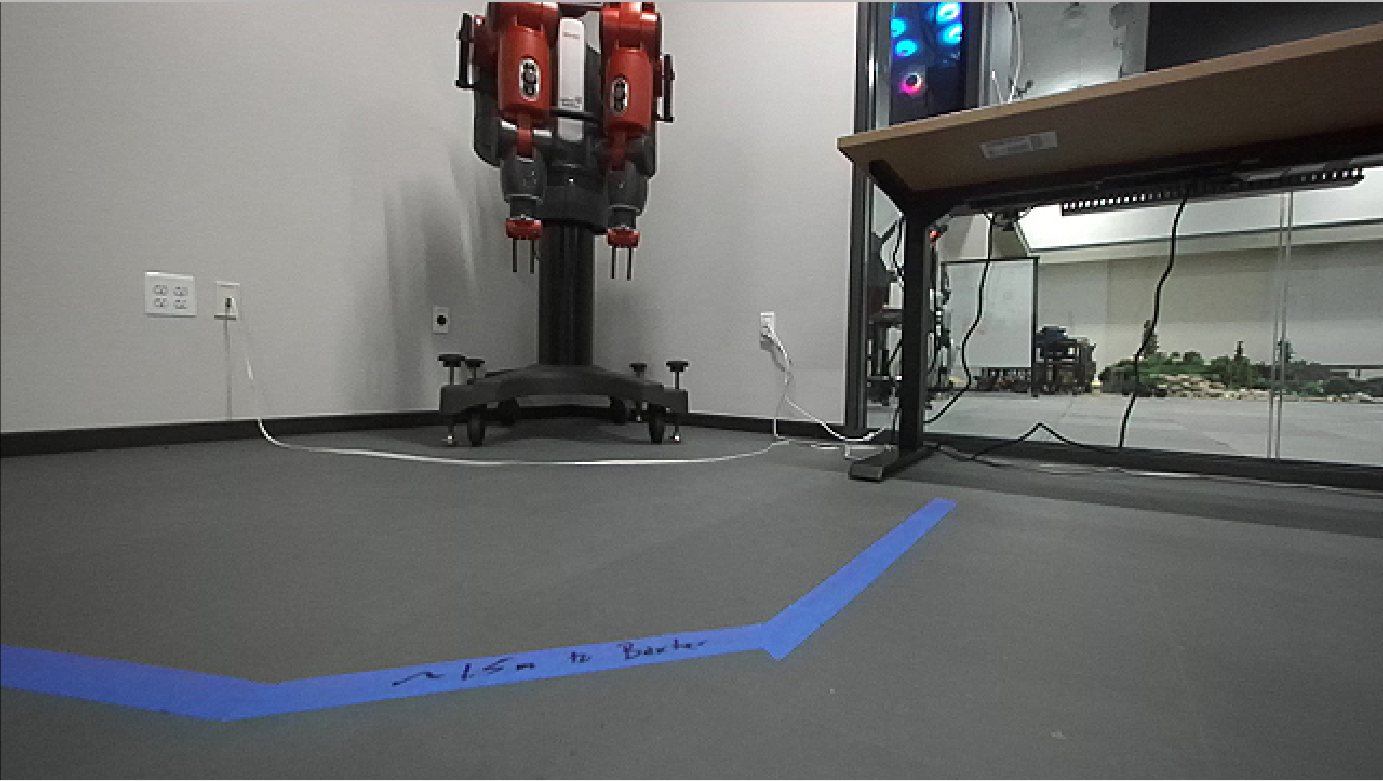} }}%
    \caption{LPIPS deviation from geometric grounding. Note that these are not generated images.}%
    \label{fig:lpips_failure}%
\end{figure}

\begin{figure}
    \centering
    \includegraphics[width=0.8\linewidth]{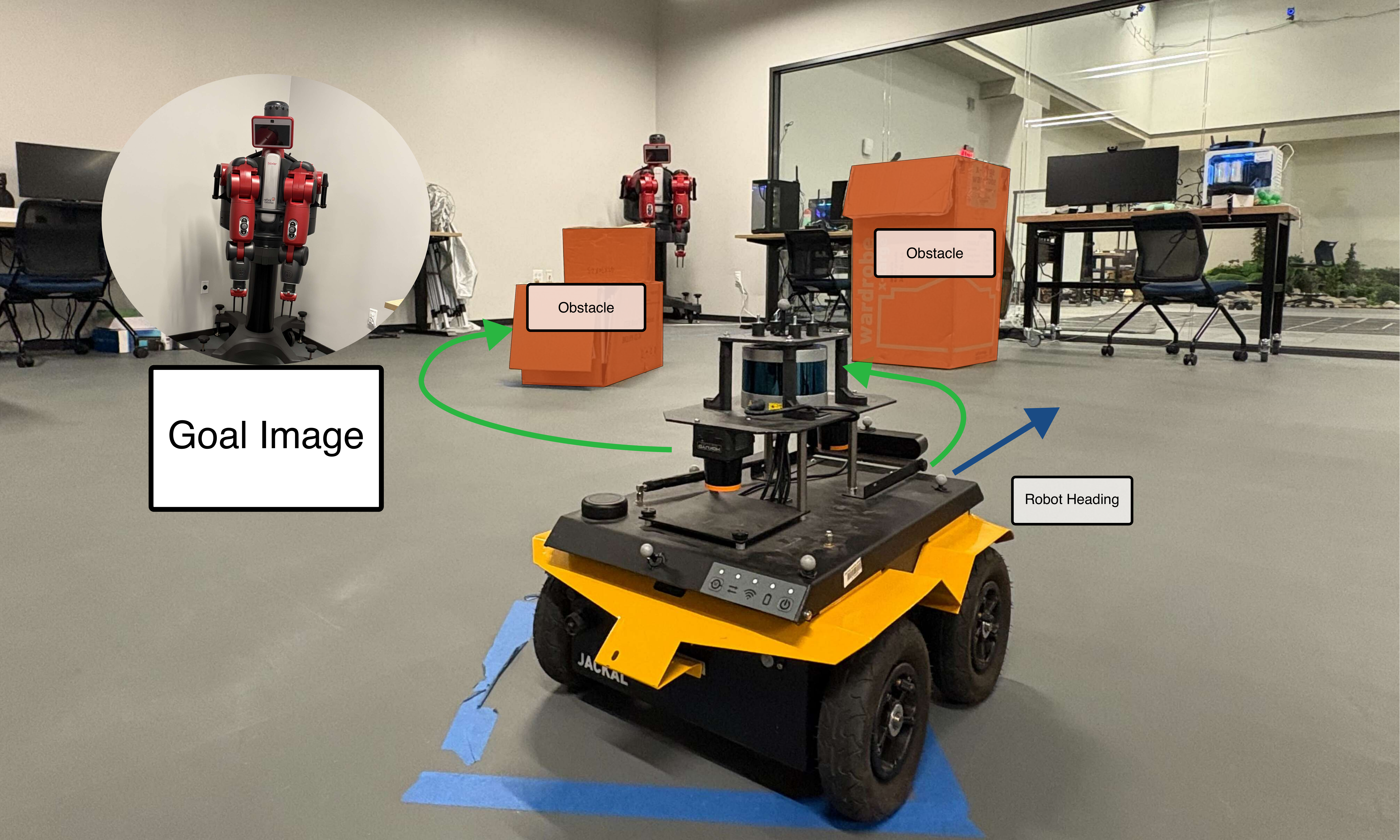}
    \caption{Physical experiment setup with obstacles.}
    \label{fig:nwm_exp_setups}
\end{figure}

\section{Architecture and Implementation Details}
\label{app:architecture}

The architecture is modularly separated into the visual-proprioceptive encoders, the unified TransformerEncoder, the discrete codebooks, and the multimodal latent flow decoders.

\textbf{Visual Encoder.} To compress high-resolution visual observations $I_t \in \mathbb{R}^{3 \times 224 \times 224}$ into spatial latents, we used the pretrained Stable Diffusion AutoencoderKL~\citep{blattmann2023stable}, which produces latents $z_i \in \mathbb{R}^{4 \times 28 \times 28}$, and kept it frozen during training. 

\textbf{Infilling Curriculum Strategy.} To ensure robustness against partial observability during deployment (\eg, missing telemetry) and to induce unconditioned exploration, we apply a learned modality infilling strategy introduced by~\cite{nazeri2025vertiformer} during training. Instead of employing a naive hard-switching dropout strategy~\citep{nazeri2025vertiformer}, we implement a cosine-annealed modality infilling scheduler. During the initial epochs, the network is trained on the fully observable joint distribution. As training progresses, the masking probability for pose and action tokens is smoothly increased following a cosine curve, preventing gradient shock and teaching the early-fusion Transformer to infer missing modalities without destabilizing the latent flow.

Over the first $E_{warmup} = 10$ epochs, we gradually decay the probability of the network receiving all modalities (the fully observable joint distribution, $p_{joint}$) from $0.5$ to $0.1$. For a given epoch $e$, the probabilities are defined as:
\begin{equation}
    p_{joint} = 0.1 + 0.4 \left[ \frac{1}{2} \left( 1 + \cos\left( \pi \min\left(1, \frac{e - 1}{E_{warmup}}\right) \right) \right) \right]
\end{equation}
The remaining probability mass is split evenly between the FKD mask state and the IKD mask state:
\begin{equation}
    p_{fkd} = p_{ikd} = \frac{1 - p_{joint}}{2}
\end{equation}
This soft-switching curriculum helps the early-fusion Transformer to learn the fully observable transition dynamics before being forced to infer missing marginals.

\textbf{Early Fusion.} Encoded actions and poses are broadcasted to a uniform length and summed to create a single, unified token stream~\citep{li2025unified}. 
To prevent receiving duplicated gradients for actions and poses due to broadcasting, which causes the model's over-sensitivity to actions and poses, we scaled down the actions and poses by the square root of their length, $\sqrt{k}$, where $k$ is the length of the broadcasted sequence. To early fuse modalities before passing them to the Transformer, \model~simply sums all the modalities to create a unified representation without introducing extra parameters, and then adds sinusoidal positional encoding~\citep{vaswani2017attention}. This simple modulation works and cuts unnecessary parameters. 

\textbf{Flow Matching Head.} To generate continuous trajectories and high-fidelity visual latents from the discrete intent vectors, \model~employs a conditional Rectified Flow Matching~\citep{liu2022flow} framework modulated via AdaLN-Zero. We deliberately tailor the representational capacity of the Flow Matching network to the specific spatial-temporal and latency constraints of each modality. 

For the high-dimensional visual world model, we utilize a full Diffusion Transformer (DiT) block~\citep{peebles2023scalable} with coordinate channels~\citep{liu2018intriguing} and a hidden dimension of 768, allowing spatial patches to continuously query each other via Multi-Head Self-Attention (MHSA) to maintain structural coherence across time. On the other hand, the DiT block for action and pose Flow Matching heads reduces the hidden dimension to 256 to achieve faster inference.

\textbf{Datasets.}
To train \model, we utilize the same heterogeneous robotic dataset corpus~\citep{shah2023gnm} as NWM. It contains 377,580 samples, broken down into 592,045,440 tokens seen during training. This corpus encompasses a diverse set of environments, including off-road~\citep{triest2022tartandrive}, indoor~\citep{hirose2023sacson}, outdoor~\citep{shah2021rapid}, and social~\citep{karnan2022socially} navigation scenarios. The dataset is structured as synchronized temporal tuples of $\{I_t, p_t, a_{t}\}$, providing the structural alignment required to jointly optimize \model's early-fusion latent manifold across visual, spatial, and kinematic domains. Unlike NWM and GNM, we \emph{do not} standardize the step size across agents by dividing the distance agents travel between frames by their maximum step size in meters, so that the model can discriminate between lower-speed robots like Clearpath Jackal and high-speed platforms such as the ATV used in TartanDrive; instead, we rely on the codebooks to capture these differences, as shown in Sec.~\ref{sec:codebook_analysis}. We filter out backward movements, following NWM and GNM.

\textbf{Training Hyperparameters.}
\label{app:training_details}
The model is trained end-to-end (with the AutoencoderKL frozen) using the AdamW optimizer. Table~\ref{tab:hydra_hyperparameters} reports the complete hyperparameter set.

\begin{table}[h!]
    \centering
    \caption{Hyperparameters and architectural details of \model. The model utilizes a Transformer backbone combined with dimensionally constrained, modality-specific VQ codebooks and Flow Matching decoders.}\label{tab:hydra_hyperparameters}
    \renewcommand{\arraystretch}{1.2}
    \begin{NiceTabular}{@{}lc@{}}
        \toprule
        \RowStyle{\bfseries} Parameter & Value \\
        \midrule

        \Block[l]{1-2}{\textbf{Training \& Optimization}} & \\
        \midrule
        Optimizer & AdamW \\
        Learning Rate & $2 \times 10^{-4}$ \\
        Weight Decay & $0.08$ \\
        Batch Size & 8 \\
        Gradient Accumulation Steps & 32 \\
        Total Effective Batch Size & 256 \\
        Epochs & 60 \\
        Gradient Clipping & 1.0 \\
        Variable Prediction Horizon & $4 \dots 12$ steps \\

        \midrule
        \Block[l]{1-2}{\textbf{Dataset Parameters}} & \\
        \midrule
        Observation Length ($H_{obs}$) & 8 steps \\
        Prediction Length ($H_{pred}$) & 12 steps \\
        Image Size & $224 \times 224$ \\
        Patch Size & $2 \times 2$ \\

        \midrule
        \Block[l]{1-2}{\textbf{Unified Transformer Architecture}} & \\
        \midrule
        Visual Backbone & SD-VAE-FT-EMA \\
        Positional Encoding & Sinusoidal 3D \\
        Transformer Layers (Encoder / Decoder) & 6 / 6 \\
        Transformer Dimension ($d_{model}$) & 768 \\
        Attention Heads & 12 \\
        Feedforward Dimension & 2048 \\
        Dropout & 0.1 \\

        \midrule
        \Block[l]{1-2}{\textbf{Vector Quantization (VQ) Bottleneck}} & \\
        \midrule
        Image Codebook Size ($\mathcal{V}_{i}$) & 2048 \\
        Action Codebook Size ($\mathcal{V}_{a}$) & 64 \\
        Pose Codebook Size ($\mathcal{V}_{p}$) & 64 \\
        Codebook Dimension & 768 \\
        Commitment Cost ($\beta$) & 0.25 \\
        EMA Decay ($\gamma$) & 0.99 \\

        \midrule
        \Block[l]{1-2}{\textbf{Flow Matching Heads}} & \\
        \midrule
        ODE Block Architecture & DiT block \\
        Hidden Dimension, Image Head & 768 \\
        Hidden Dimension, Pose/Action Heads & 256 \\
        Sampling Steps, Image Head (NFEs) & 50 \\
        Sampling Steps, Pose/Action Heads & 10 \\
        ODE Source Distribution & $\mathcal{N}(0, \mathbf{I})$ \\
        Deterministic Loss Weight & 0.1 \\

        \bottomrule
    \end{NiceTabular}
\end{table}

\textbf{Variable Target Horizon.} While the maximum prediction horizon is fixed to 3 seconds ($H=12$ steps at 4Hz), we utilize a variable-length temporal regularization strategy during training. For each batch, the target rollout length is uniformly sampled between $1$ and $3$ seconds. This prevents the model from overfitting to a rigid temporal boundary, ensuring robust transition dynamics at arbitrary planning horizons during physical deployment.

\section{Architectural Ablations}
\label{sec:arch_ablations}

To isolate the specific contributions of \model's architectural components, we perform a series of ablation studies focusing on the discrete intent bottleneck, context compression, and inference memory constraints.

\subsection{Projection Capacity and Codebook Utilization}
\label{sec:ab:utilization}
To empirically validate the necessity of the strict linear information bottleneck preceding the Flow Matching solver, we ablated the projection layer architecture. We compared our standard linear projection against a higher-capacity 2-layer MLP with GeLU activation. We evaluated the models at 237K training steps, analyzing both codebook perplexity (the entropy of discrete code utilization) and the cosine similarity of the multi-task gradients.

While both architectures achieved $100\%$ code usage, the linear projection yielded a significantly higher image codebook perplexity ($608.36$) compared to the MLP ($465.42$). This indicates that the non-linear MLP possesses enough capacity to lazily route information, resulting in an unstructured, low-entropy utilization of the visual manifold. Conversely, the strict linear projection forces the network to uniformly distribute representations across the available codebook.

\begin{figure}
    \centering
    \includegraphics[width=0.9\linewidth]{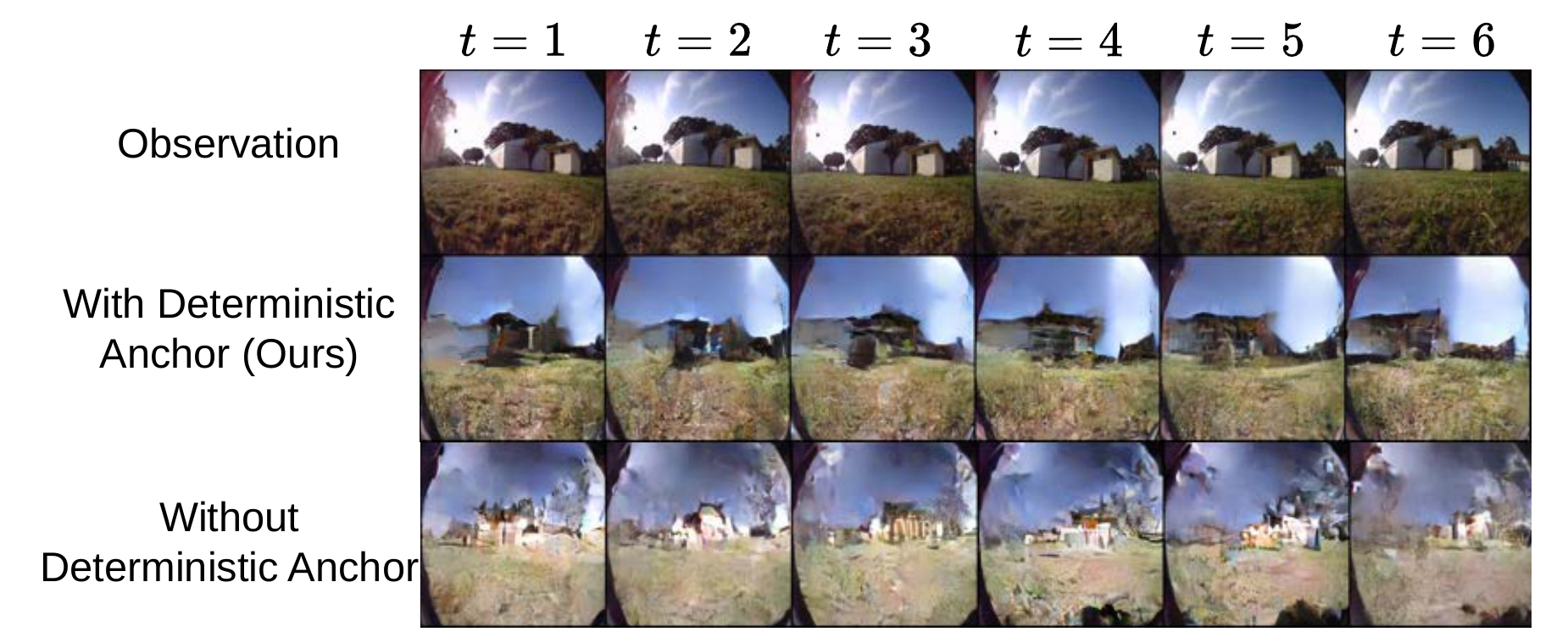}
    \caption{Early Training Dynamics (2.1K Steps). A comparison between the model's early-stage generation, both with (middle) and without (bottom) the deterministic anchor head. While the bottom row suffers from spatial smearing, the model utilizing the deterministic head shows a better ability to construct structural details, like the building's edges against the sky, even after very limited training.}
    \label{fig:nodet}
\end{figure}

\subsection{Deterministic Decoder Heads}
\label{app:deterministic_heads}
Besides the projection capacity, to empirically validate the necessity of this deterministic projection preceding the VQ codebooks, we visualized the early training dynamics (at 2.1K steps) with and without this auxiliary head. As shown in Figure~\ref{fig:nodet}, passing randomly initialized codebooks directly into the continuous flow-matching solver causes severe representational instability. Conversely, incorporating the deterministic head provides an immediate stabilizing prior. By grounding the codebooks with a standard MSE loss, the network is forced to maintain spatial coherence early in the training process, preventing the downstream generative ODE from diverging and accelerating the formation of structural scene details.

\section{Latent Planning Algorithm Details}
\label{app:stochastic_planning}

\subsection{Two Sources of Stochasticity}
\label{app:two_noise_sources}

Generating diverse candidate trajectories from a discrete VQ latent space requires randomizing two different things, at two different stages of the search.

\textbf{Why codebook-level noise alone is not enough.} A natural first attempt is to compute the Transformer's intent tokens once, then let each of the $B$ candidates independently sample a codebook entry from the resulting logits. This produces candidates that differ only in \emph{which nearby codebook vector each one snaps to}, and because the codebook is trained so that nearby vectors are semantically similar, swapping the 1st-nearest for the 3rd-nearest vector rarely changes the decoded trajectory's shape in any meaningful way. This gives quantization jitter around a single underlying idea (\eg, ``swerve left'' vs.\ ``swerve left, very slightly differently''), never qualitatively different intents (\eg, ``swerve left'' vs.\ ``swerve right'').

\textbf{Site 1: Gaussian noise on the continuous intent query.} To obtain genuinely different trajectory shapes, the input to the Transformer decoder itself must change. Since the Transformer, not the codebook, decides the high-level shape of the candidate we therefore perturb the continuous intent query independently for each candidate before it is passed to the decoder:
\begin{equation}
    \tilde{\epsilon}_b[t] = \alpha_{sm} \cdot \epsilon_b[t] + (1 - \alpha_{sm}) \cdot \tilde{\epsilon}_b[t-1], \qquad \epsilon_b \sim \mathcal{N}(0, I),
\end{equation}
smoothed across the horizon via an Exponential Moving Average (EMA) with coefficient $\alpha_{sm}$ so that a perturbation at one timestep biases neighboring timesteps in a similar direction, producing a smooth alternate path (\eg, ``curve further right the whole way'') rather than step-to-step jitter that would be physically implausible for a robot to execute. Each candidate's noisy query, $q_b^{(i)} = P_{intent}^{(i-1)} + \sigma^{(i)} \tilde{\epsilon}_b$, is then passed through a fresh Transformer forward pass, yielding $B$ genuinely different sets of intent tokens rather than $B$ discretizations of one shared set. The noise scale $\sigma^{(i)}$ shrinks geometrically across iterations, giving the search a coarse-to-fine, CEM-like character: broad exploration early, local refinement late.

\textbf{Site 2: Top-$K$ Gumbel-max discretization.} Each candidate's (now distinct) continuous intent tokens must still be snapped onto valid codebook vectors before decoding. We do this stochastically, restricted to the $K$ nearest vectors by cosine similarity, rather than with a hard nearest-neighbor snap, adding a second, finer layer of exploration around each of Site 1's hypotheses while the top-$K$ restriction keeps candidates from drifting to codebook entries that are cosine-far, \ie, off-manifold and physically implausible (Sec.~\ref{sec:codebook_analysis}). Like Site 1, the raw Gumbel noise driving this discretization is itself temporally smoothed via its own EMA, with its own coefficient $\alpha_{gum}$, independent of Site 1's $\alpha_{sm}$: without this, independent per-timestep Gumbel draws would let each candidate's discretization jitter step to step even when its underlying continuous trajectory (from Site 1) is smooth. This is detailed in the remainder of this section.

Both sites are necessary and serve different purposes: Site 1 diversifies \emph{what the model considers}; Site 2 makes each consideration's discretization robust rather than collapsed onto a single nearest neighbor, while remaining temporally coherent in its own right. Algorithm~\ref{alg:energy_intent_search} uses both simultaneously, each independently smoothed and annealed.

\begin{algorithm}[ht]
\caption{Discrete Latent Planning (\model)}
\label{alg:energy_intent_search}
\begin{algorithmic}[1]
\Require Observation history $\mathcal{H}_t$, goal pose $P_{goal}$, visual goal $z_{goal}$, horizon $T$, candidates $B$, iterations $N$
\State $h_{ctx} \gets \text{UnifiedEncoder}(\mathcal{H}_t)$ \Comment{encode history once}
\State $P^{(0)} \gets$ unbiased expert prior pose sequence, broadcast to $B$ candidates
\State $J_{best} \gets \infty$, $c_{best} \gets \emptyset$
\For{$i = 0$ to $N-1$}
    \If{$i > 0$}
        \State $P^{(i)} \gets P^{(i-1)} +$ decaying smoothed Gaussian noise, per candidate \Comment{explore pose space}
    \EndIf
    \State $z_{e} \gets \text{TransformerDecoder}(h_{ctx}, P^{(i)})$
    \State $c \gets$ discretize $z_e$ via Top-$K$-restricted, annealed Gumbel-Softmax sampling
    \State $\hat{P}, \hat{z}_{img} \gets$ decode $c$ to pose and visual codes
    \State $J \gets \text{KPC}(\hat{P}, \hat{z}_{img}, P_{goal}, z_{goal})$ \Comment{Eq.~\eqref{eq:kpc}, per candidate}
    \State $b^* \gets \arg\min_b J_b$
    \If{$J_{b^*} < J_{best}$}
        \State $J_{best} \gets J_{b^*}$; $c_{best} \gets c_{b^*}$
    \EndIf
    \State $P^{(i+1)} \gets$ momentum update toward $\hat{P}_{b^*}$, broadcast to $B$ candidates \Comment{shift search region}
\EndFor
\State \Return $\text{FlowDecoder}(\text{Lookup}(c_{best}, \mathcal{V}_p))$ \Comment{high-fidelity continuous execution}
\end{algorithmic}
\end{algorithm}

Note that $J_{best}$ and $c_{best}$ track the globally best candidate across \emph{all} $N$ iterations.

\textbf{Standard Gumbel-Max Trick.}
Let $s_{t, c}$ denote the unnormalized logit predicted by the Transformer decoder for timestep $t \in \{1, \dots, T\}$ and codebook entry $c \in \mathcal{V}_p$. Sampling from the Categorical distribution $p(c_t) \propto \exp(s_{t, c} / \rho)$ with temperature $\rho$ is equivalent to the Gumbel-Max trick: $$ c_t = \arg\max_{c \in \mathcal{V}} \left( \frac{s_{t, c}}{\rho} + g_{t, c} \right) $$ where $g_{t, c} \sim \text{Gumbel}(0, 1)$ is i.i.d.\ noise sampled via $g_{t, c} = -\log(-\log(u_{t, c}))$ with $u_{t, c} \sim \mathcal{U}(0, 1)$.

\textbf{Iteration-Dependent Annealing Schedule.}
Since the search is an iterative optimizer running for $N$ depth iterations, we anneal the Site 2 temperature $\rho$ and Site 2's own smoothing coefficient $\alpha_{gum}$ at each iteration $i$. Site 1's smoothing coefficient $\alpha_{sm}$ is, by contrast, a fixed constant and is not annealed; only its noise \emph{scale} $\sigma^{(i)}$ shrinks across iterations.

\emph{Simulated Annealing (Coarse-to-Fine Search):} We exponentially decay the Site 2 temperature to transition from broad topological exploration to greedy exploitation: $$ \rho^{(i)} = \rho_0 \cdot 0.1^{\,i} $$ where $\rho_0$ is the base temperature.

\emph{Annealing the Gumbel Smoothing Factor:} We linearly increase $\alpha_{gum}$ as iterations progress: $$ \alpha_{gum}^{(i)} = \min\big(0.4 + 0.15 \cdot i,\; 1.0\big) $$ During early iterations, $\alpha_{gum}^{(i)}$ is small, generating highly correlated Gumbel noise for large-scale, smooth exploration; in late iterations, $\alpha_{gum}^{(i)} \to 1.0$, allowing fine-grained, near-independent micro-adjustments to individual timesteps along the elite trajectory.

\textbf{Top-$K$ Kinematic Masking.}
Even with temporally correlated exploration, the planner might occasionally sample outlier codes that violate kinematic constraints. We strictly bound the sampling space by projecting a hard constraint mask over the logits. Let $\mathcal{T}_t$ be the set containing the indices of the $K$ largest logits at timestep $t$. We define the mask $\mathbf{M} \in \{0, -\infty\}$ as: $$ M_{t, c} = \begin{cases}
0 & \text{if } c \in \mathcal{T}_t \\
-\infty & \text{otherwise}
\end{cases}$$
This formulation ensures that the sampled latent intent remains within the physically plausible manifold learned by \model~(enforced by $\mathbf{M}$), while the Site 1 perturbation ensures that the sampled intents exhibit the temporal coherence necessary for stable robotic control. Table~\ref{tab:dlp_hyperparameters} shows the DLP hyperparameters used during experiments.

\begin{table}[h!]
    \centering
    \caption{DLP Hyperparameters.}\label{tab:dlp_hyperparameters}
    \renewcommand{\arraystretch}{1.2}
    \begin{NiceTabular}{@{}lccccccc@{}}
        \toprule

        \textbf{$\lambda_{\text{img}}$} & \textbf{$\lambda_{\text{prior}}$} & \textbf{$\lambda_{\text{ent}}$} & \textbf{$\lambda_{\text{vq}}$} & \textbf{$\gamma$} & \textbf{\# Candidates} & \textbf{Iterations} & \textbf{$top_k$ samples} \\
        \midrule

        0.02  & 0.5  & 0.5 & 1.0 & 0.85 & 18 & 3 & 10 \\

        \bottomrule
    \end{NiceTabular}
\end{table}

\subsection{Classifier-Free Guidance for Navigation Trajectory Generation}
\label{app:cfg}
\paragraph{Vocabulary-based Maneuver Candidate Generation.}
To circumvent the OOD hallucination risk of latent
extrapolation, we restrict the search space to the valid pose codebook
$\mathcal{V}_p$. We observe that the codebook itself encodes a diverse maneuver
vocabulary (Fig.~\ref{fig:codebook_coverage}): its entries, decoded as intents, span endpoint headings covering both turn signs. Given the
mean-pooled unconditional prior $\bar{z}_{\emptyset} = \frac{1}{K}\sum_{k=1}^{K} z_{\emptyset}^{(k)}$, where $K$ is an arbitrary number, we query the model unconditionally to get a more stable intent because of the randomness in FM heads. Then, we construct a batch of $N$ candidate intents in a single forward pass as:
\begin{equation}
  \mathcal{C} =
  \underbrace{\{\bar{z}_{\emptyset}\}}_{\text{expert default}}
  \;\cup\;
  \underbrace{\{ \mathbf{e}_c \}_{c \in \mathcal{V}_p}}_{\text{pure maneuvers}}
  \;\cup\;
  \underbrace{\{ \bar{z}_{\emptyset} + \lambda_j (\mathbf{e}_{c_j} - \bar{z}_{\emptyset}) \}}_{\text{convex mixes}},
  \quad \lambda_j \sim \mathcal{U}[0.25, 1],
  \label{eq:candidates}
\end{equation}
where $\mathbf{e}_c$ denotes codebook embedding $c$, broadcast as a constant
intent over the horizon, and the candidate budget is split evenly between pure
codebook entries and convex mixes that interpolate between the scene-conditioned
prior and the vocabulary. Importantly, every candidate is re-quantized through the
VQ bottleneck before decoding, so even the mixes are snapped
back onto the discrete manifold; by construction only valid codes are ever
decoded. Where stochastic diversity is desired, we add small Gaussian
perturbations drawn \emph{once per candidate and timestep}, shared across the $H$
intent slots and smoothed over time by an exponential moving average. Sharing the
noise across slots is essential. 

\paragraph{Deterministic Intent Decoding.}
Candidates are decoded with the deterministic pose head $D_{\text{pose}}$, App.~\ref{app:deterministic_heads}), rather than the
flow-matching head. The deterministic head is pointwise stable, which makes
candidate scores comparable across the batch, and it is trained on the intents it receives here. The stochastic ODE decoder, in contrast,
exhibits high per-seed variance for maneuver-code candidates that lie far from the prior mode, which would render candidate comparison difficult. The same decode path is then used to execute the winner, so selection and execution are identical by construction. Since the deterministic decode is a single matrix multiplication per
candidate, scoring $N$ hypotheses is cheaper than a single flow-matching
integration, preserving the low latency required for real-time control.

\subsection{Sampling: Candidate Path Guidance}\label{app:sampling}
When the global planner supplies a reference path $\mathcal{G} = \{g_1, \dots, g_s\}$, we therefore score each candidate with a truncated Chamfer distance to $\mathcal{G}$. Let
$\ell_t = \sum_{i \leq t} \| p_i - p_{i-1} \|_2$ 
be the candidate's arc length and $L_{\mathcal{G}}$ the total arc length of the reference. We retain only the prefix $\mathcal{K} = \{ t : \ell_t \le L_{\mathcal{G}} \}$ and compute:
\begin{equation}
  d_{\text{path}}(\tau) =
  \frac{1}{|\mathcal{K}|} \sum_{t \in \mathcal{K}} \min_{s} \| \tau_t - g_s \|_2
  \;-\; w_{\text{prog}} \cdot \rho_{\text{reward}}(\tau),
  \qquad
  \rho_{\text{reward}}(\tau) = \frac{\arg\min_{s} \| \tau_T - g_s \|_2}{S - 1},
  \label{eq:chamfer}
\end{equation}
where each candidate point is matched to its \emph{nearest} reference point, so candidates are not penalized for executing the path at a different speed than the reference, and $\rho_{\text{reward}}(\tau)$ is a progress reward measuring how far along $\mathcal{G}$ the candidate reaches. The path cost is blended with the candidate's deviation from the unconditional expert trajectory $p^{\text{prior}}_{1:T}$:
\begin{equation}
  \mathcal{J}_{\text{track}}(\tau) =
  \omega \cdot d_{\text{path}}(\tau)
  \;+\; \max(1 - \omega,\, 0) \cdot \underbrace{\frac{1}{T} \sum_{t} \| p_t - p_t^{\text{prior}} \|_2}_{\text{prior deviation}},
  \label{eq:pathcost}
\end{equation}
where $\omega{=}1$ yields strict path following and $\omega{<}1$ increasingly trusts the obstacle-avoiding prior. While $\omega{>}1$ is permitted, it reduces the second term to zero, amplifying the path signal. The selected intent is decoded to continuous actions and executed directly.

\textbf{Computational Trade-offs.} By restricting candidate generation exclusively to the unbiased expert prior, every sampled trajectory lies on the learned manifold. While evaluating visual predictive entropy or manifold rejection over these candidates (as done in Sec.~\ref{sec:planning}) would provide obstacle avoidance, we explicitly omit the image codebook and decoder from both local sampling approaches. This omission is a strategic pruning choice designed to maintain low inference latency, relying entirely on the inherent safety of the expert before handling local obstacle avoidance. The primary limitation of this method occurs in highly constrained environments (\eg, sharp U-turns), where the unconditional prior may assign near-zero probability to the necessary maneuver, relying on the diversity of sampling methods.

\section{Experimental Setup and Evaluation Protocols}\label{app:experiments_setup}

In this section, we detail the physical deployment configurations used to evaluate the discrete latent planning (DLP) and local path following capabilities of \model.

\subsection{Goal-Directed Planning Evaluation}

To evaluate the computational efficiency and planning capabilities of our DLP formulation, we deploy \model~across two distinct physical embodiments: a wheeled differential-drive robot (Clearpath Jackal) and a quadrupedal robot (Boston Dynamics Spot). Given that unmapped, long-horizon planning remains a largely unsolved challenge in autonomous navigation, our experimental scope is intentionally bounded to close-range, complex local navigation tasks.

We define two primary evaluation scenarios:
\begin{itemize}
	\item \textbf{Variable-Initialization Point Navigation:} The robot must plan and execute a collision-free trajectory to a Cartesian coordinate approximately $d = 8.0$ meters away. To evaluate robustness to initial state perturbations, the robot is initialized in 5 varying starting orientations, sampled uniformly from $\theta_{init} \in [-40^{\circ}, +40^{\circ}]$ relative to the goal vector. This is evaluated in both unobstructed spaces and partially occluded environments.

	\item \textbf{Constrained Corridor Maneuvering:} The robot must execute a high-curvature turn (approximately $90^{\circ}$) within a narrow corridor, testing the model's ability to respect tight geometric constraints during planning.

\end{itemize}

A trial succeeds if the robot navigates within 0.5 meters of the goal without collision. We define trial success $\mathcal{S}$ for a trajectory $\tau = \{x_0, x_1, \dots, x_T\}$ with respect to a goal position $x_{goal} \in \mathbb{R}^2$ as:$$\mathcal{S}(\tau) = \mathds{1} \left( \left( \min_{t \in [0, T]} \vert{}\vert{} x_t - x_{goal} \vert{}\vert{}_2 \leq 0.5 \text{ m} \right) \land \left( \sum_{t=0}^{T} \mathcal{C}(x_t) = 0 \right) \right)$$where $\mathcal{C}(x_t)$ is a binary collision indicator function at time step $t$.

\textbf{Compute Architecture Constraints:}
Due to the computation demand of NWM, on-board edge computation is currently bottlenecked by memory bandwidth and FLOP limits. Consequently, all physical inference is offloaded to a high-compute workstation equipped with an NVIDIA A5000 GPU (24GB VRAM).

\subsection{Local Path Following Settings}

To isolate and evaluate the robustness of the downstream local path following policy, we construct three specialized physical scenarios representing distinct environmental priors:

\begin{itemize}
	\item \textbf{Cluttered Goal-Conditioned Traversal (Obstructed Goal):} The robot is tasked with navigating a narrow corridor seeded with static, unmapped obstacles. This setting evaluates the policy's capacity for on-the-fly obstacle avoidance and reactive re-planning while maintaining progress toward a line-of-sight goal.

	\item \textbf{Multi-Waypoint Navigation (Long Distance):} The robot must execute a long-horizon navigation task. This includes navigating multiple intersections within a building's corridor network, culminating in a room entry and a stop adjacent to a designated target. This evaluates the model's temporal consistency over extended rollouts.

	\item \textbf{Strict Boundary Adherence (Indoor Corridor):} The robot operates in a narrow, feature-sparse indoor corridor. The primary objective is to maintain a safe margin from the walls. This setting tests the model's geometric boundary perception in constrained space.
\end{itemize}

Performance is measured using strict geometric bounds (\eg, $L_2$ norm to the goal) and boolean collision constraints.

%% file: contents/comparison_table.tex
\begin{table}[h!]
\centering
\caption{Comparison with visually-grounded robot navigation baselines. \textbf{I}: Image, \textbf{P}: Pose, \textbf{A}: Action.}
\label{tab:comparison}
\scalebox{0.7}{
\begin{NiceTabular}{lccccc}
\toprule
\RowStyle{\bfseries} Method & Obs. Input & Goal Input & Cond. Input & Output & Nav. Policy \\
\midrule
GNM~\citep{shah2023gnm} & I & I & I & P & Reactive \\
VINT~\citep{shah2023vint} & I & I & I & P & Reactive \\
NOMAD~\citep{sridhar2023nomad} & I & I & I & P & Reactive \\
VertiFormer~\citep{nazeri2025vertiformer} & I, A, P & I, P & A or P & I, A, P & Planning \\
NWM~\citep{bar2025navigation} & I & I & P & I & Planning \\
\midrule
\rowcolor{LightBlueBg} \textbf{\model} & I, A, P & I, P & A or P & I, A, P & Planning \\
\bottomrule
\end{NiceTabular}
}
\vspace{2mm}
\end{table}

%% file: contents/dataset_comparison_table.tex
\begin{table}[h]

\centering
\caption{Offline trajectory forecasting evaluation on a held-out subset across three navigation datasets (SCAND, TartanDrive, and SACSoN). \model~achieves competitive policy accuracy while retaining the planning capability that reactive baselines lack.}
\label{tab:offline_eval}
\footnotesize
\setlength{\tabcolsep}{3pt}
\begin{tabular}{ll cccccccc}
\toprule
\textbf{Model} & \textbf{Params} & \multicolumn{2}{c}{SCAND} & \multicolumn{2}{c}{TartanDrive} & \multicolumn{2}{c}{SACSoN} & \multicolumn{2}{c}{Mean} \\
\cmidrule(lr){3-4} \cmidrule(lr){5-6} \cmidrule(lr){7-8} \cmidrule(lr){9-10}
& & ADE $\downarrow$ & FDE $\downarrow$ & ADE $\downarrow$ & FDE $\downarrow$ & ADE $\downarrow$ & FDE $\downarrow$ & ADE $\downarrow$ & FDE $\downarrow$ \\
\midrule
GNM & 8.6M & 0.16 & 0.27 & 0.67 & 1.12 & 0.13 & 0.21 & 0.32 & 0.53 \\
ViNT & 29.6M & 0.15 & 0.25 & 0.80 & 1.33 & 0.16 & 0.26 & 0.37 & 0.61 \\
NoMaD & 19.0M & 0.24 & 0.43 & 1.311 & 2.35 & 0.22 & 0.39 & 0.59 & 1.06 \\
VertiFormer & 27.11M & \textbf{0.10} & \textbf{0.18} & \textbf{0.43} & \textbf{0.80} & \textbf{0.10} & \textbf{0.19} & \textbf{0.21} & \textbf{0.39} \\
\midrule
\rowcolor{LightBlueBg} \textbf{\model} & 143.29M & 0.12 & 0.20 & 0.48 & 0.85 & 0.12 & 0.21 & 0.24 & 0.42 \\
\bottomrule
\end{tabular}
\vspace{-1em}
\end{table}